\documentclass[twocolumn,natbib]{svjour3}

\usepackage{hyperref}
\usepackage{graphicx}
\usepackage{amsmath,amssymb, array, multirow, caption}
\usepackage{color, lettrine}
\usepackage{xspace}
\usepackage{soul}
\usepackage{algorithm, algpseudocode}
\usepackage{comment}

\newtheorem{mydef}{Definition}

\def\CNN{CNN}
\def\CNNs{CNN~}
\def\pcknspace{Patch-CKN~}
\def\pckn{Patch-CKN}

\def\kmone{{k\text{--}1}}
\def\lmone{{l\text{--}1}}
\def\x{{ x}}
\def\y{{x'}}
\def\w{{w}}
\def\z{{v}}
\def\Real{{\mathbb R}}
\newcommand{\mikodat}{Mikolajczyk et al's dataset}
\newcommand{\mikodatspace}{Mikolajczyk et al's dataset~}

\titlerunning{Convolutional Patch Representations for Image Retrieval: an Unsupervised Approach}
\title{Convolutional Patch Representations for Image Retrieval: an Unsupervised Approach}

\authorrunning{Mattis Paulin, Julien Mairal, Matthijs Douze, Zaid Harchaoui, Florent Perronnin, Cordelia Schmid}
\author{Mattis Paulin \and
  Julien Mairal \and 
  Matthijs Douze \and 
  Zaid Harchaoui \and 
  Florent Perronnin \and
  Cordelia Schmid}

\begin{document}\sloppy

\institute{Thoth team, Inria Grenoble Rhone-Alpes, Laboratoire Jean Kuntzmann, CNRS, Univ. Grenoble Alpes, France.}

\maketitle 

\begin{abstract}
Convolutional neural networks (CNNs) have recently received a lot of attention
due to their ability to model local stationary structures in natural images in
a multi-scale fashion, when learning all model parameters with supervision.
While excellent performance was achieved for image classification when large amounts of labeled
visual data are available, their success for unsupervised tasks such as image
retrieval has been moderate so far. 

Our paper focuses on this latter setting and explores several methods
for learning patch descriptors without supervision with application
to matching and instance-level retrieval.  To that effect, we propose
a new family of 
convolutional descriptors for patch representation, based on the
recently introduced convolutional kernel networks.  We show that our
descriptor, named \pckn, performs better than SIFT as well as other
convolutional networks learned by artificially introducing supervision
and is significantly faster to train. To demonstrate its
effectiveness, we perform an extensive evaluation on standard
benchmarks for patch and image retrieval where we obtain
state-of-the-art results.  We also introduce a new dataset called
RomePatches, which allows to simultaneously study descriptor performance for patch and image retrieval.
 
\keywords{Low-level image description, Instance-level retrieval, Convolutional Neural Networks.}
\end{abstract}

\begin{figure*}[t]
\begin{center}
  \includegraphics[width=0.9\textwidth]{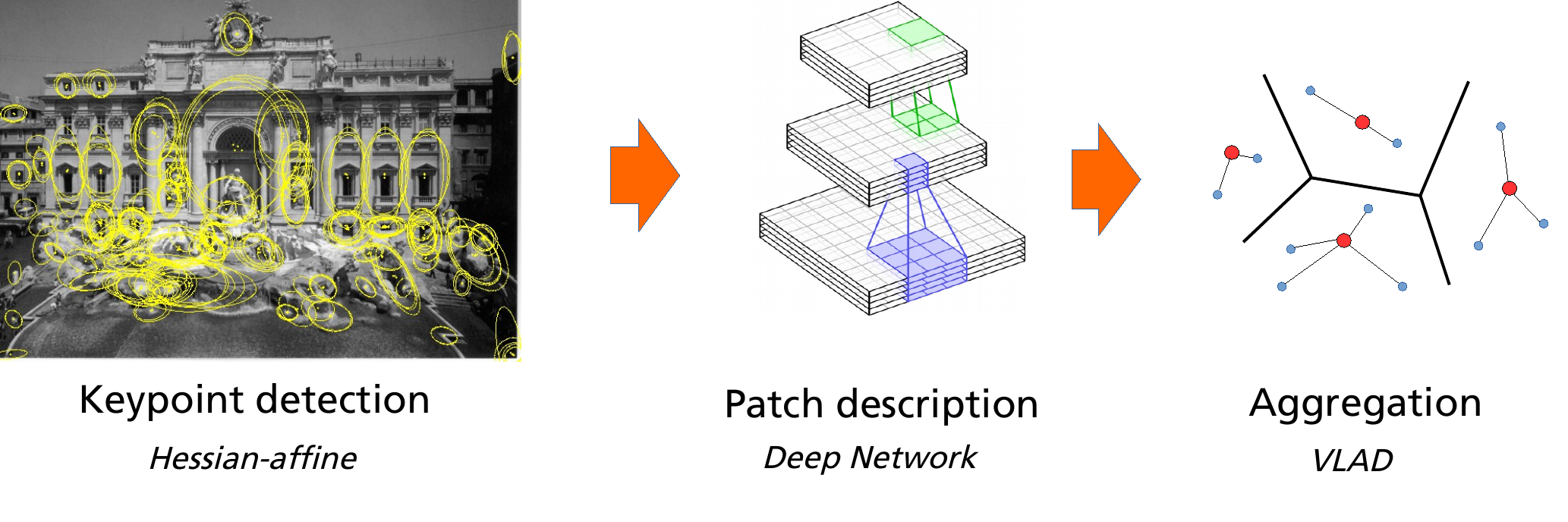}\vspace{-0.5cm}
  \caption{\label{fig:pipeline} Proposed image retrieval
    pipeline. Interest points are extracted with the Hessian-affine
    detector (left), encoded in descriptor space (middle), and
    aggregated into a compact representation using
    VLAD-pooling (right). The main focus of the paper lies in the choice of the patch-level
    visual descriptor.}
\end{center}
\end{figure*}

\section{Introduction}

\lettrine[lines=2]{T}{his} paper explores convolutional architectures as robust visual
descriptors for image patches and evaluates them in the context
of patch and image retrieval. We explore several levels of supervision for
training such networks, ranging from fully supervised to unsupervised.
In this context, requiring supervision may seem unusual since data for
retrieval tasks typically does not come with labels. Convolutional
Neural Networks (CNNs) have achieved state-of-the-art in many other computer
vision tasks, but require abundant labels to learn their
parameters. For this reason, previous work with CNN architectures on 
image retrieval have focused on using global~\citep{babenko2014neural}
or aggregated
local~\citep{razavian2014cnn,gong2014multi,ng2015exploiting,babenko2015aggregating,tolias2015particular}
CNN descriptors that were learned on an unrelated classification
task. To improve the performance of these transferred features,
\citet{babenko2014neural} showed that fine-tuning global descriptors on
a dataset of landmarks results in improvements on retrieval datasets that
contain buildings. It is unclear, however, if the lower levels of a convolutional
architecture, appropriate for a local description, will be impacted by such a global fine-tuning~\citep{yosinsky2014transfer}. 
Recent approaches have, thus, attempted to discriminatively learn low-level convolutional
descriptors, either by enforcing a certain level of invariance through explicit
transformations~\citep{fischer2014descriptor} or by training with a patch-matching
dataset~\citep{zagoruyko2015learning,simoserra2015discriminative}. In
all cases, the link between the supervised classification objective
and image retrieval is artificial, which motivates us to investigate
the performance of new unsupervised learning techniques. 

To do so, we propose an unsupervised patch descriptor based on
Convolutional Kernel Networks (CKNs)\citep{mairal2014convolutional}. This
required to turn the CKN proof  of concept
of~\citet{mairal2014convolutional} into a descriptor with
state-of-the-art performance on large-scale benchmarks. This paper
introduces  significant improvements of the original model, algorithm,
and implementation, as well as adapting the approach to image
retrieval. Our conclusion is that supervision might not be necessary
to train convolutional networks for image and patch retrieval, since
our unsupervised descriptor achieves the best performance on several
standard benchmarks.

One originality of our work is also to \emph{jointly} evaluate our models on
the problems of \emph{patch} and~\emph{image} retrieval.
Most works that study patch
representations~\citep{brown2011discriminative,
winder2009picking,zagoruyko2015learning,fischer2014descriptor}, do so
in the context of patch retrieval only, and do not test whether 
conclusions also generalize to image retrieval (typically after an
aggregation step). In fact, the correlation between 
the two evaluation methods (patch- and image-level) is not clear beforehand, which
motivated us to design a new dataset to answer this question. We call this
dataset RomePatches; it consists of views of several
locations in Rome~\citep{li2010location}, for which a sparse
groundtruth of patch matches is obtained through 3D
reconstruction. This results in a dataset for patch and image retrieval, which
enables us to quantify the performance of patch descriptors for both tasks.

To evaluate the descriptor performance, we adopt the following
pipeline (see Fig.~\ref{fig:pipeline}), described in detail in section \ref{sec:pipeline}. We
use the popular Hessian-Affine detector of 
\cite{mikolajczyk2004scale}, which has been shown to give
state-of-the-art results~\citep{tinne2007}. The regions around these
points are encoded with the convolutional descriptors proposed in this work. We aggregate local
features with VLAD-pooling~\citep{jegou2010aggregating} on the patch
descriptors to build an approximate matching technique. VLAD pooling has been shown
to be better than Bag-of-Word, and to give similar performance to
Fisher Vectors~\citep{perronnin2007fisher}, another popular technique
for image retrieval~\citep{jegou2012aggregating}.

A preliminary version of this article has appeared
in~\citep{paulin2015local}. Here we extend the related work
(Section~\ref{sec:related}) and describe in detail the convolutional
kernel network (CKN), in particular its reformulation which leads
to a fast learning algorithm (Section~\ref{sec:ckns}). 
We also add a number of experiments (see Section~\ref{sec:experiments}). 
We compare to two additional supervised local CNN descriptors.
The first one is based on the network of~\cite{fischer2014descriptor}, but trained on our
RomePatches dataset. The second one is trained with the Siamese
architecture of~\cite{zagoruyko2015learning}.  
We also compare our approach to an AlexNet network fine-tuned on
the Landmarks dataset. Furthermore, we provide an in-depth study of
PCA compression and whitening, standard post-processing 
steps in image retrieval, which further improve the quality of
our results. We also show that our method can be applied to
dense patches instead of Hessian-Affine, which improves performance
for some of the benchmarks.

The remainder of the paper is organized as follows. We discuss previous
work that is most relevant to our approach in Section~\ref{sec:related}. We describe
the framework for convolutional 
descriptors and convolutional kernel networks in
Sections~\ref{sec:convdescs} and~\ref{sec:ckns}. We introduce the 
RomePatches dataset as well as standard benchmarks for 
patch and image retrieval in Section~\ref{sec:datasets}. 
Section~\ref{sec:experiments} describes experimental results.

\section{\label{sec:related}Related Work} 


In this section we first review the state of the art for patch
description and then present deep learning approaches for image and
patch retrieval.  For deep patch descriptors, we first present 
supervised and, then,   unsupervised approaches.

\subsection{Patch descriptors}

A patch is a image region extracted from an image. Patches can either
be extracted densely or at interest points. 
The most popular patch descriptor is
SIFT~\citep{lowe2004distinctive}, which showed state-of-the-art
performance\cite{mikolajczyk2005performance} for patch matching.  
It can be viewed as a  three-layer CNN, the first layer computing
gradient histograms using convolutions, the second, fully-connected,
weighting the gradients with a Gaussian, and the third  pooling across
a 4x4 grid.  
Local descriptors that improve SIFT include SURF~\citep{bay2006surf},
BRIEF~\citep{brief2010} and LIOP~\citep{liop2011}.  Recently,
\cite{dong2015domain} build on SIFT using local pooling on scale   
and location to get state-of-the-art performance in patch retrieval. 

All these descriptors are hand-crafted and their relatively small
numbers of parameters have been optimized by grid-search. 
When the number of parameters to be set is large,
such an approach is unfeasible and the optimal parametrization needs
to be learned from data.  
  
A number of approaches learn patch descriptors without relying on deep
learning. Most of them use a strong
supervision. \cite{brown2011discriminative} (see also 
\cite{winder2009picking}) design a matching dataset 
based on 3D models of landmarks and use it to train a descriptor
consisting of several existing parts, including 
SIFT, GLOH~\citep{mikolajczyk2005performance} and Daisy
\citep{tola2010daisy}.
\cite{philbin2010descriptor} learn a Mahalanobis metric for SIFT
descriptors to compensate for the binarization error, with excellent
results in instance-level retrieval. \cite{simonyan2014learning}
propose the ``Pooling Regions'' descriptor and learn its parameters,
as well as a linear projection using stochastic optimization. 
Their learning objective can be cast as a convex optimization problem, 
which is not the case for classical convolutional networks.

An exception that departs from this strongly supervised setting is~\citep{bo2010kernel} which presents a match-kernel interpretation of SIFT,
and a family of kernel descriptors whose parameters are learned in an unsupervised fashion.
The \pcknspace we introduce generalizes kernel descriptors; the proposed procedure
for computing an explicit feature embedding is faster and simpler.

\subsection{Deep learning for image retrieval} 

If a \CNNs is trained on a sufficiently large labeled
set such as ImageNet~\citep{deng2009imagenet},
its intermediate layers can be used as image descriptors
for a wide variety of tasks including image
retrieval~\citep{babenko2014neural,razavian2014cnn}.
The output of one of the fully-connected layers is often
chosen because it is compact, usually 4,096-dim. However, global
\CNNs descriptors lack geometric invariance~\citep{gong2014multi}, and
produce results below the state of the art for instance-level image
retrieval. 

In~\citep{razavian2014cnn,gong2014multi}, \CNNs
responses at different scales and positions are extracted.
We proceed similarly, yet we replace the (coarse) dense grid with a patch detector.
There are important differences between~\citep{razavian2014cnn,gong2014multi} and our work. 
While they use the penultimate layer as patch descriptor, 
we show in our experiments that we can get improved results with 
preceding layers, that are cheaper to compute and require smaller input patches.
Closely related is the work of \cite{ng2015exploiting} which uses VLAD
pooling on top of very deep \CNNs feature maps, at multiple scales with
good performance on Holidays and Oxford. Their approach is similar to
the one of \cite{gong2014multi}, but faster as it factorizes
computation using whole-image convolutions. Building on this,
\cite{tolias2015particular} uses an improved aggregation method
compared to VLAD, that leverages the structure of convolutional
feature maps.  

\cite{babenko2014neural} use a single global \CNNs
descriptor for instance-level image retrieval and fine-tune the
descriptor on an external landmark dataset.
We experiment with their fine-tuned network and show improvement also
with lower levels on the Oxford dataset. CKN descriptors still
outperform this approach.
Finally, \citep{wang2014learning} proposes a Siamese architecture 
to train image retrieval descriptors but do not report results on standard retrieval benchmarks.

\subsection{Deep patch descriptors} 
Recently~\citep{long2014corresp,fischer2014descriptor,simoserra2015discriminative,
zagoruyko2015learning} outperform SIFT for patch matching or patch classification.
These approaches use different levels of supervision to train a \CNN.
\citep{long2014corresp} learn their patch CNNs using category labels
of ImageNet.  \citep{fischer2014descriptor} creates surrogate classes where each class corresponds
to a patch and distorted versions of this patch.
Matching and non-matching pairs are used  in \citep{simoserra2015discriminative,
  zagoruyko2015learning}. 
There are two key differences between those works and ours.
First, they focus on patch-level metrics, instead of actual image retrieval.
Second, and more importantly, while all these approaches require some kind of supervision,
we show that our \pcknspace  yields competitive performance
in both patch matching and image retrieval without any supervision.

\subsection{Unsupervised learning for deep representations} 

To avoid costly annotation,  many works leverage unsupervised
information to learn deep representations.  
Unsupervised learning can be used to initialize network weights, as in
\cite{erhan2009difficulty,erhan2010does}. 
Methods that directly use unsupervised weights include domain
transfer~\citep{donahue2014decaf} and
k-means~\citep{coates2012learning}. Most recently, some works have
looked into using temporal coherence as supervision~\citep{goroshin2015learning,
goroshin2015unsupervised}. Closely related to our work,
\cite{agrawal2015learning} propose to train a network by learning the affine
transformation between synchronized image pairs for which camera parameters are available.
Similarly, \cite{jayaraman2015learning} uses a
training objective that enforces for sequences of images that derive
from the same ego-motion to behave similarly in the feature space. 
While these two works focus on a weakly supervised setting, 
we focus on a fully unsupervised one.

\section{\label{sec:convdescs} Local Convolutional Descriptors} 


In this section, we briefly review notations related to CNNs and
the possible learning approaches.

\subsection{Convolutional Neural Networks}

In this work, we use convolutional features to encode patches
extracted from an image. We call
convolutional descriptor any feature representation $f$ that decomposes in a
\emph{multi-layer} fashion as:
\begin{equation}\label{eq:convdescs}
  f(x) = \gamma_K(\sigma_K(W_K\dots \gamma_2(\sigma_2(W_2\gamma_1(\sigma_1(W_1x))\dots)),
\end{equation}
where $x$ is an input patch represented as a vector, the $W_k$'s are matrices corresponding to
linear operations, the $\sigma_k$'s are pointwise non-linear functions,
e.g., sigmoids or rectified linear units, and the functions $\gamma_k$
perform a downsampling operation called ``feature pooling''. 
Each composition $\gamma_k(\sigma_k(W_k \bullet ))$ is called a
``\emph{layer}'' and the intermediate representations of~$x$, between each layer, are called
``\emph{maps}''. A map can be represented as pixels organized on a spatial grid, with a
multidimensional representation for each pixel.
Borrowing a classical terminology from neuroscience, it is also common to call
``\emph{receptive field}'' the set of pixels from the input patch~$x$ that may
influence a particular pixel value from a higher-layer map. 
In traditional convolutional neural networks, the $W_k$ matrices have a
particular structure corresponding to spatial convolutions performed by small
square filters, which will need to be learned. In the case where there is no
such structure, the layer is called ``fully-connected''. 

The hyper-parameters of a convolutional architecture lie in the choice
of non-linearities $\sigma_k$, type of pooling $\gamma_k$, in the structure of
the matrices $W_k$ (notably the size and number of filters) as well as in the
number of layers. 

The only parameters that are learned in an automated fashion are
usually the filters, corresponding to the entries of the matrices $W_k$. In
this paper we investigate the following ways of learning:
 (i) encoding local descriptors with a \CNNs that has been
trained for an unrelated classification task (Sec. \ref{sec:category}), (ii) using a \CNNs that
has been trained for a classification problem that can be directly linked
to the target task (e.g. buildings, see Sec. \ref{sec:category}), (iii) devising a surrogate
classification problem to enforce invariance (Sec. \ref{sec:surrogate}), (iv) directly learning
the weights using patch-level groundtruth (Sec. \ref{sec:patchgt}) or (v) using unsupervised
learning, such as convolutional kernel networks, which we present in  Section~\ref{sec:ckns}. 

\subsection{Learning Supervised Convolutional Descriptors}

The traditional way of learning the weights ${\mathcal{W} = (W_1, W_2,\dots,W_K)}$ in (\ref{eq:convdescs})
consists in using a training set ${\mathcal{X} = (x_1,x_2,\dots,x_n)}$ of
examples, equipped with labels
${\mathcal{Y} = (y_1,y_2,\dots,y_n)}$, choose a loss
function ${\ell(\mathcal{X},\mathcal{Y},\mathcal{W})}$ and minimize
it over $\mathcal{W}$ using stochastic gradient optimization and
back-propagation \citep{lecun1989handwritten,bottou2012stochastic}.
The choice of examples, labelings and loss functional leads to
different weights.

\begin{figure}
   \centering
   \includegraphics{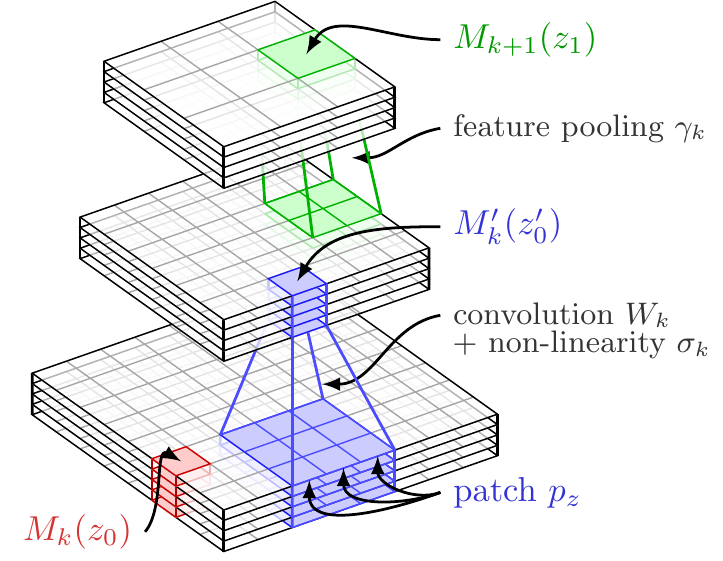}
   \label{subfig:cnn}
   \caption{Typical organization of two successive layers of a CNN. The
     spatial map~$M'_{k}$ is obtained from~$M_{k}$ by convolution and pointwise
     non-linearity, and the top layer~$M_{k+1}$ is obtained from $M'_{k}$ by a
     downsampling operation called feature pooling. Usual CNNs use
     ReLU non-linearities and max-pooling
     \citep{krizhevsky2012imagenet}, while CKNs rely on exponentials and
     Gaussian pooling \citep{mairal2014convolutional}.} 
\label{fig:sketch} 
\end{figure}

\subsubsection{\label{sec:category}Learning with category labels} A now classical
CNN architecture is AlexNet~\citep{krizhevsky2012imagenet}. AlexNet consists of 8 layers: the first five are convolutional layers
and the last ones are fully connected. 

In this case, the training examples are images that have been hand-labeled into
$C$ classes such as ``bird'' or ``cow'' and the loss function is the
softmax loss:

\begin{equation}\label{eq:softmax}
   \ell(\mathcal{X},\mathcal{Y},\mathcal{W}) = \sum_{i=1}^n
  \log\sum_{j=1}^C \exp\left(M_K^{(i)}[j] - M_K^{(i)}[y_i]\right).
\end{equation}
In Eq.~(\ref{eq:softmax}) and throughout the paper, $M_k^{(i)}$ is
the output of the $k$-th layer ($k\in\{1,\dots,K\}$) of the network
applied to example $x_i$. The $[j]$ notation corresponds to the $j$-th
element of the map.

Even though the network is designed to process images of size $224
\times 224$, each neuron of a map has a ``receptive field'', see the 
``coverage'' column in Table~\ref{tab:patch_results} from
Section~\ref{sec:experiments}. Using an image of the size of the
receptive field produces a 1x1 map that we can use as a
low dimensional patch descriptor. To ensure a fair comparison between all
approaches, we rescale the fixed-size input patches so that they fit the
required input of each network.

We explore two different sets of labellings for AlexNet: the first
one, which we call AlexNet-ImageNet, is learned on the training set of ILSVRC 2012 ($C=1000$), as in the original paper~\citep{krizhevsky2012imagenet}.
This set of weights is popular in off-the-shelf
convolutional features, even though the initial task is unrelated to
the target image retrieval application. Following
\citet{babenko2014neural}, we also fine-tune the same network on the
Landmarks dataset, to introduce semantic
information into the network that is more related to the target
task. The resulting network is called AlexNet-Landmarks.

\subsubsection{Learning from surrogate labels\label{sec:surrogate}}

Most CNNs such as AlexNet augment the dataset with 
jittered versions  of training data to learn the filters~$W_k$
in~(\ref{eq:convdescs}). \cite{dosovitskiy2014discriminative,fischer2014descriptor} use virtual 
patches, obtained as transformations of randomly extracted ones to design a
classification problem related to patch retrieval.
For a set of patches~$\mathcal{P}$, and a set a transformations
$\mathcal{T}$, the dataset consists of all $\tau(p),~(\tau,
p)\in\mathcal{T}\times\mathcal{P}$. Transformed versions of the same 
patch share the same label, thus defining surrogate
classes. Similarily to the previous setup, the network use 
softmax loss (\ref{eq:softmax}). 

In this paper, we evaluate this strategy by using the same network, 
called PhilippNet, as in \cite{fischer2014descriptor}.
The network has three convolutional and one fully connected layers, takes
as input 64x64 patches, and produces a 512-dimensional output. 

\subsection{Learning with patch-level groundtruth\label{sec:patchgt}}

When patch-level labels are available, obtained by
manual annotation or 3D-reconstruction \citep{winder2009picking}, it is
possible to directly learn a similarity measure as well as a feature
representation. The simplest way to do so, is to replace the virtual
patches in the architecture of
\cite{dosovitskiy2014discriminative,fischer2014descriptor} described
in the previous section with labeled patches of RomeTrain. We call
this version ``FisherNet-Rome''.  

It can also be achieved using a Siamese network 
\citep{chopra2005learning}, i.e.\ a CNN which takes as input the two patches to
compare, and where the objective function enforces that the output
descriptors' similarity should reproduce the ground-truth similarity
between patches.

Optimization can be conducted with either a metric-learning cost
\citep{simoserra2015discriminative}:

%

\begin{equation}
\ell(\mathcal{X},\mathcal{Y},W_K) = \sum_{i=1}^n\sum_{j=1}^n
C(i,j,\|M^{(i)}_K - M^{(j)}_K\|)
\end{equation}
with 
\begin{equation}
C(i,j,d) = \left\{
\begin{array}{ll}
d & \mathrm{if}~y_i = y_j \\
\max(0, 1-d) & \mathrm{otherwise}
\end{array}
\right.
\end{equation}
or as a binary classification problem
(``match''/``not-match'') with a softmax loss as in
eq. (\ref{eq:softmax}) \citep{zbontar2014computing,
  zagoruyko2015learning}. For those experiments, we use the parameters
of the siamese networks of \cite{zagoruyko2015learning}, available
online\footnote{\url{https://github.com/szagoruyko/cvpr15deepcompare}}. Following
their convention, we refer to these architectures as ``DeepCompare''.

\section{\label{sec:ckns}Convolutional Kernel Networks}
 
In this paper, the unsupervised learning strategy for learning convolutional 
networks is based on the convolutional kernel networks (CKNs) of
\citet{mairal2014convolutional}. Similar to CNNs, these networks have a
multi-layer structure with convolutional pooling and nonlinear
operations at every layer. Instead of learning filters by optimizing a loss
function, say for classification, they are trained layerwise to approximate a
particular nonlinear kernel, and therefore require no labeled data. 

The presentation of CKNs is divided into three stages: (i) introduction of the
abstract model based on kernels (Sections~\ref{subsec:ckn1}, \ref{subsec:ckn2}, and
\ref{subsec:ckn3}); (ii) approximation scheme and concrete implementation (Sections~\ref{subsec:ckn4}, \ref{subsec:ckn5}, and \ref{sec:inputs});
(iii) optimization (Section~\ref{subsec:ckn6}).


\subsection{A Single-Layer Convolutional Kernel for Images}\label{subsec:ckn1}
The basic component of CKNs is a match kernel that encodes a similarity
between a pair of images $(M,M')$ of size $m\times m \times d$ pixels, which
are assumed to be square. The integer $d$ represents the number of
channels, say~3 for RGB images.
Note that when applied to image retrieval, these images~$M,M'$ correspond
to regions -- patches -- extracted from an
image. We omit this fact for simplicity since this presentation of
CKNs is independent of the image retrieval task.

We denote by
$\Omega$ the set of pixel locations, which is of size $|\Omega| = m
\times m$, and choose a patch size $e \times e$. Then, we denote
by $P_z$ (resp. $P'_{z'}$) the $e\times e\times d$ patch of $M$
(resp. $M'$) at location $z\in\Omega$ (resp. $z'\in\Omega$). 
Then, the single-layer match kernel is defined as follows:
\begin{mydef}{Single-Layer Convolutional Kernel.}
  \begin{equation}\label{eq:ckn-single}
     K(M,M') = \sum_{z,z'\in\Omega} e^{-\frac{1}{2\beta^2}\|z-z'\|^2} \kappa(P_z, P'_{z'}),
  \end{equation}
  with 
  \begin{equation}\label{eq:ckn-kappa1}
     \kappa(P, P') = \|P\|\|P'\| e^{-\frac{1}{2\alpha^2}\|\tilde P - \tilde
    P'\|^2},
\end{equation}
where $\alpha$ and $\beta$ are two kernel hyperparameters, 
${\Vert \cdot \Vert}$ denotes the usual $\ell_2$ norm, and $\tilde{P}$
and~$\tilde{P}'$ are $\ell_2$-normalized versions of the
patches~$P$ and $P'$.
\end{mydef}

$K$ is called a \emph{convolutional kernel}; it can be interpreted as a
match-kernel that compares all pairs of patches from~$M$ and~$M'$ with a
nonlinear kernel~$\kappa$, weighted by a Gaussian term that decreases with
their relative distance. The kernel compares indeed all locations in $M$ with 
all locations in $M'$. It depends notably on the parameter $\alpha$, which
controls the nonlinearity of the Gaussian kernel comparing two normalized patches~$\tilde{P}$ and~$\tilde{P}'$, 
and on~$\beta$, which controls the size of the neighborhood in which a patch is matched with another one. 
In practice, the comparison of two patches that have very different
locations~$z$ and~$z'$ will be negligible in the sum~(\ref{eq:ckn-single}) when~$\beta$ is
small enough.
Hence, the parameter $\beta$ allows us to control the local shift-invariance of
the kernel.  

\subsection{From Kernels to Infinite-Dimensional Feature Maps}\label{subsec:ckn2}
Designing a positive definite kernel on data is equivalent to defining a
mapping of the data to a Hilbert space, called reproducing kernel Hilbert space
(RKHS), where the kernel is an inner product~\citep{rkhs-book:2007}; exploiting this mapping
is sometimes referred to as the ``kernel trick''~\citep{scholkopf2002learning}.
In this section, we will show how the kernel~(\ref{eq:ckn-single}) may be used
to generalize the concept of ``feature maps'' from the traditional neural
network literature to kernels and Hilbert spaces.\footnote{Note that in the kernel
literature, ``feature map'' denotes the mapping between data points and their
representation in a reproducing kernel Hilbert space (RKHS).
Here, feature maps refer to spatial maps representing local image
characteristics at every location, as usual in the neural network
literature~\cite{lecun1998}.}  The kernel $K$ is indeed positive
definite (see the appendix of~\citealt{mairal2014convolutional}) and thus it
will suits our needs.

Basically, feature maps from convolutional neural networks are spatial maps
where every location carries a finite-dimensional vector representing
information from a local neighborhood in the input image.
Generalizing this concept in an infinite-dimensional context is relatively
straightforward with the following definition:
%
\begin{mydef} Let $\mathcal{H}$ be a Hilbert space. The set of feature
   maps is the set of applications $\varphi:\Omega\rightarrow \mathcal{H}$.  
\end{mydef}
Given an image~$M$, it is now easy to build such a feature map. For instance, consider the
nonlinear kernel for patches~$\kappa$ defined in Eq.~(\ref{eq:ckn-kappa1}).
According to the Aronzsjan theorem, there exists a Hilbert space~$\mathcal{H_{\kappa}}$ and 
a mapping $\phi_{\kappa}$ such that for two image patches~$P,P'$ -- which may come from
different images or not --, $\kappa(P,P')=\langle
\phi_{\kappa}(P), \phi_{\kappa}(P') \rangle_{\mathcal H_{\kappa}}$.
As a result, we may use this mapping to define a feature map~$\varphi_{M}:
\Omega \to {\mathcal H}_{\kappa}$ for image~$M$ such that $\varphi_M(z) = \phi_{\kappa}(P_z)$,
where~$P_z$ is the patch from~$M$ centered at location~$z$.  The first property
of feature maps from classical CNNs would be satisfied: at every
location, the map carries information from a local neighborhood from the input
image~$M$.

We will see in the next subsection how to build sequences of feature maps in a
multilayer fashion, with invariant properties that are missing from the simple example
we have just described.

%
%
%

\subsection{From Single-Layer to Multi-Layer Kernels}\label{subsec:ckn3}
\begin{figure}
   \centering
   \includegraphics[width=0.45\textwidth]{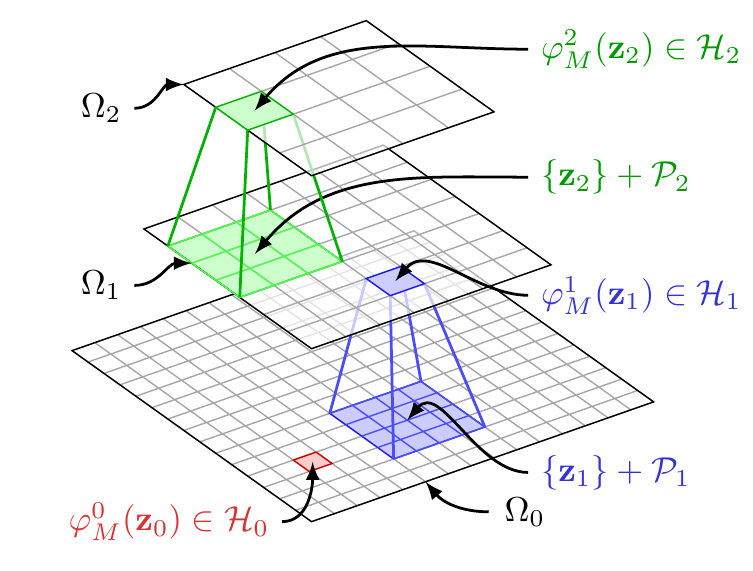}
   \caption{Construction of the sequence of feature maps~$\varphi_M^1,\ldots,\varphi_M^k$ for an input image~$M$. The point $\varphi_M^k(z)$ for $z$ in~$\Omega_k$ represents the mapping in~${\mathcal H}_k$ of a patch from~$\varphi_M^{\kmone}$ centered at~$z$. Figure adapted from~\citet{mairal2014convolutional}.} \label{fig:ckn1}
\end{figure}

We now show how to build a sequence of feature maps~$\varphi^1_{M}$,
\ldots, $\varphi^k_{M}$ for an input image~$M$ initially represented as a finite-dimensional
map~$\varphi^0_{M} : \Omega_0 \to {\mathbb R}^{p_0}$, where~$\Omega_0$ is the
set of pixel locations in~$M$ and~$p_0$ is the number of channels.
The choice of initial map~$\varphi_M^0$ is important since it will be the input
of our algorithms; it is thus discussed in Section~\ref{sec:inputs}. Here, we
assume that we have already made this choice, and we explain how to build a
map~$\varphi_M^{k}: \Omega_{k} \to {\mathcal H}_{k}$ from a previous
map~$\varphi_M^{\kmone}: \Omega_\kmone \to {\mathcal H}_{\kmone}$.  
Specifically, our goal is to design~$\varphi_M^{k}$ such that
\begin{itemize}
   \item[(i)] $\varphi_M^{k}(z)$ for~$z$ in~$\Omega_{k}$ carries information from a local neighborhood from~$\varphi_M^\kmone$ centered at location~$z$;
   \item[(ii)] the map $\varphi_M^{k}$ is ``more invariant'' than~$\varphi_M^\kmone$. 
\end{itemize}
These two properties can be obtained by defining a positive definite
kernel~$K_{k}$ on patches from~$\varphi_{M}^\kmone$. Denoting by~${\mathcal
H}_k$ its RKHS, we may call $\varphi_M^k(z)$ the mapping to~${\mathcal
H}_k$ of a patch from~$\varphi_M^{\kmone}$ centered at~$z$.
The construction is illustrated in Figure~\ref{fig:ckn1}.
 
Concretely, we choose a patch shape~${\mathcal P}_{k}$, which is a set of
coordinates centered at zero along with a
set of pixel locations~$\Omega_{k}$ such that for all~$z$ in~$\Omega_{k}$
and~$u$ in~${\mathcal P}_k$, the location~$z+u$ is in~$\Omega_{\kmone}$. Then,
the kernel~$K_k$ for comparing two patches from~$\varphi_M^\kmone$
and~$\varphi_{M'}^\kmone$ at respective locations~$z,z'$ in~$\Omega_k$ is defined as
\begin{equation}
   \sum_{u,u' \in {\mathcal P}_k}  e^{-\frac{1}{2\beta_k^2}\|u-u'\|^2} \kappa_k(\varphi_M^\kmone(u+z), \varphi_M^\kmone(u'+z')), \label{eq:cknpatch}
\end{equation}
where 
\begin{displaymath}
   \kappa_k(\varphi,\varphi') = \|\varphi\|_{{\mathcal H}_\kmone}  \|\varphi'\|_{{\mathcal H}_\kmone} e^{-\frac{1}{2\alpha_k^2}\|\tilde{\varphi} -\tilde{\varphi}' \|^2_{{\mathcal H}_\kmone}},   
\end{displaymath}
for all~$\varphi,\varphi'$ in~${\mathcal H}_{\kmone}$, where 
$\tilde{\varphi}$ (resp. $\tilde{\varphi}'$) are normalized---that is, $\tilde{\varphi} = (1/\|\varphi\|_{{\mathcal H}_\kmone})\varphi$ if~$\varphi=0$
and~$0$ otherwise.
This kernel is similar to the convolutional kernel for images already
introduced in~(\ref{eq:ckn-single}),
except that it operates on infinite-dimensional feature maps.  It involves two
parameters~$\alpha_k$, $\beta_k$ to control the amount of invariance of the 
kernel. Then, by definition,~$\varphi_M^k: \Omega_k
\to {\mathcal H}_k$ is the mapping such that the value (\ref{eq:cknpatch}) is
equal to the inner product  $\langle
\varphi_M^k(z),\varphi_{M'}^k(z')\rangle_{{\mathcal H}_k}$.

This framework yields a sequence of infinite-dimensional image
representations but requires finite-dimensional approximations to be used
in practice. Among different approximate kernel embeddings
techniques~\citep{williams2001using,Bach:Jordan:2002,Rahimi:Recht:2008,PerronninSL10,VedaldiZ12}, 
we will introduce a data-driven approach that exploits a simple expansion of
the Gaussian kernel, and which provides a new way of learning convolutional
neural networks without supervision.

\subsection{Approximation of the Gaussian Kernel}\label{subsec:ckn4}
Specifically, the previous approach relies on an approximation scheme for the
Gaussian kernel, which is plugged in the convolutional
kernels~(\ref{eq:cknpatch}) at every layer; this scheme  
requires learning some weights that will be interpreted as the parameters
of a CNN in the final pipeline (see Section~\ref{subsec:ckn5}).


More precisely, for all~$\x$ and~$\y$ in~$\Real^q$, and~$\alpha >
0$, the Gaussian kernel $e^{-\frac{1}{2\alpha^2}\|\x-\y\|_2^2}$ can be shown to be equal to
\begin{equation}
   \left(\frac{2}{\pi \alpha^2}\right)^{\frac{q}{2}} \int_{\z \in \Real^q} e^{-\frac{1}{\alpha^2}\|\x-\z\|_2^2}e^{-\frac{1}{\alpha^2}\|\y-\z\|_2^2} d\z.\label{eq:rbf}
\end{equation}
Furthermore, when the
vectors~$\x$ and~$\x'$ are on the sphere---that is, have unit $\ell_2$-norm, we
also have 
\begin{equation}
   e^{-\frac{1}{2\alpha^2}\|\x-\y\|_2^2} = {\mathbb E}_{\z \sim p(\z)}[s(\z^\top \x) s(\z^\top \y)],\label{eq:integral0}
\end{equation}
where $s$ is a nonlinear function such that $s(u) \propto
e^{-\frac{1}{\alpha^2} + \frac{2 u}{\alpha^2}}$ and $p(\z)$ is the density of the multivariate normal distribution $
{\mathcal N}(0,(\alpha^2/4) {\mathbf I})$.
Then, different strategies may be used to approximate the expectation
by a finite weighted sum:
\begin{equation}
   e^{-\frac{1}{2\alpha^2}\|\x-\y\|_2^2} \approx \frac{1}{p}\sum_{j=1}^p \eta_j s(\z_j^\top \x) s(\z_j^\top \y),\label{eq:integral}
\end{equation}       
which can be further simplified, after appropriate changes of variables,
\begin{equation}
   e^{-\frac{1}{2\alpha^2}\|\x-\y\|_2^2} \approx \sum_{j=1}^p e^{\w_j^\top \x + b_j} e^{\w_j^\top \y + b_j},\label{eq:integral2}
\end{equation}
for some sets of parameters~$\w_j$ in~$\Real^p$ and~$b_j$ in~$\Real$, $j=1,\ldots,p$, which need to be learned.
The approximation leads to the kernel approximations $\langle \psi(\x),\psi(\y)
\rangle$ where~$\psi(\x) = [e^{\w_j^\top \x + b_j}]_{j=1}^p$, which may be interpreted
as the output of a one-layer neural network with~$p$ neurons and exponential
nonlinear functions.

The change of variable that we have introduced yields a simpler formulation than the original formulation of
\citet{mairal2014convolutional}. Given a set of training pairs of normalized 
signals~$(\x_1,\x_1'),\ldots,(\x_n,\x_n')$ in~$\Real^q$, the weights~$\w_j$ and
scalars~$b_j$ may now be obtained by minimizing
            %
            %
%
%
\begin{equation}\label{eq:cknoptimization}
    \min_{W,b} \quad \sum_{i=1}^n \left[e^{\frac{\Vert x_i -x'_i\Vert^2}{2\alpha^2}}
       - \sum_{j=1}^p e^{w_j^\top  x_i + b_j}
    e^{w_j^\top x'_i + b_j} \right] ^2,
 \end{equation}
which is a non-convex optimization problem. How we address it will be detailed in
Section~\ref{sec:sgdopt}.
%
%
\subsection{\label{sec:cknapprox}Back to Finite-Dimensional Feature Maps}\label{subsec:ckn5}
Convolutional kernel networks use the previous approximation scheme of the Gaussian
kernel to build finite-dimensional image representations $M_1: \Omega_1 \to \Real^{p_1}, M_2: \Omega_2 \to \Real^{p_2}, \ldots,
M_k: \Omega_k \to \Real^{p_k}$ of an input image~$M$ with the following properties:
\begin{itemize}
   \item[(i)] There exists a patch size $e_k \times e_k$ such that a patch $P_{l,z}$ of $M_l$ at
      location~$z$---which is formally a vector of size $e_k^2 p_{\kmone}$---provides a finite-dimensional approximation of the kernel
      map~$\varphi_M^l(z)$. In other words, given another patch~$P'_{l,z'}$ from a map~$M'_l$, we have
      $\langle \varphi_M^l(z), \varphi_M^l(z)\rangle_{{\mathcal H}_l}  \approx \langle P_{l,z} , P'_{l,z'} \rangle$.\footnote{Note that to be more rigorous, the maps $M_l$ need to be slightly larger in spatial size than~$\varphi_M^l$ since otherwise a patch~$P_{l,z}$ at location~$z$ from~$\Omega_l$ may take pixel values outside of~$\Omega_l$. We omit this fact for simplicity.}
   \item[(ii)] Computing a map $M_l$ from~$M_{\lmone}$ involves convolution with learned filters and linear feature pooling with Gaussian weights.
\end{itemize}
Specifically, the learning and encoding algorithms are presented in
Algorithms~\ref{alg:learning} and~\ref{alg:encoding}, respectively.
The resulting procedure is relatively simple and admits two interpretations.
\begin{itemize}
   \item The first one is to consider CKNs as an approximation of the
infinite-dimensional feature maps~$\varphi_M^1,\ldots,\varphi_M^k$ presented in the
previous sections~\citep[see][for more details about the approximation principles]{mairal2014convolutional}.
\item The second one is to see CKNs as particular types of convolutional
   neural networks with contrast-normalization. Unlike traditional CNNs,
   filters and nonlinearities are learned to approximate the Gaussian kernel on
   patches from layer~$\kmone$.
\end{itemize}
With both interpretations, 
this representation induces a change of paradigm in unsupervised learning with
neural networks, where the network is not trained to reconstruct input signals, 
but where its non-linearities are derived from a kernel point of view.


\begin{algorithm}[hbtp]
  \begin{algorithmic}
     \State \textsc{Hyper-parameters:} Kernel parameter $\alpha_k$, patch size $e_k \times e_k$, number of filters~$p_k$.
     \State \textsc{Input model:} A CKN trained up to layer~$\kmone$. 
     \State \textsc{Input data:} A set of training images. 
     \State \textsc{Algorithm:}
     \begin{itemize}
        \item Encode the input images using the CKN up to layer~$\kmone$ by using Algorithm~\ref{alg:encoding};
        \item Extract randomly $n$ pairs of patches~$(P_i,P'_i)$ from the maps obtained at layer $\kmone$;
        \item Normalize the patches to make then unit-norm;
        \item Learn the model parameters by minimizing~(\ref{eq:cknoptimization}), with $(x_i,x_i')=(P_i,P_i')$ for all~$i=1,\ldots,n$;
     \end{itemize}
    \State\textsc{Output:} Weight matrix~$W_k$ in~$\Real^{p_{\kmone}e_k^2 \times p_k}$ and $b_k$ in~$\Real^{p_k}$.
  \end{algorithmic}
  \caption{\label{alg:learning} Training layer~$k$ of a CKN.}
  \end{algorithm}
\begin{algorithm}[hbtp]
  \begin{algorithmic}
     \State \textsc{Hyper-parameters:} Kernel parameter $\beta_k$;
     \State \textsc{Input model:} CKN parameters learned from layer~$1$ to~$k$ by using Algorithm~\ref{alg:learning};
     \State \textsc{Input data:} A map $M_\kmone: \Omega_{k-1} \to \Real^{p_{\kmone}}$; 
     \State \textsc{Algorithm:}
     \begin{itemize}
        \item Extract patches~$\{P_{k,z}\}_{z \in \Omega_{\kmone}}$ of size~$e_k \times e_k$ from the input map $M_\kmone$;
        \item Compute contrast-normalized patches
           \begin{displaymath}
              \tilde{P}_{k,z} = \frac{1}{\|P_{k,z}\|}P_{k,z} ~~\text{if}~~P_{k,z} \neq 0~~~\text{and}~~0~~\text{otherwise}. 
           \end{displaymath}
        \item Produce an intermediate map $\tilde{M}_k: \Omega_{\kmone} \to \Real^{p_k}$ with linear operations followed by non-linearity: 
           \begin{equation}\label{eq:tmp_map}
              \tilde{M}_k(z) = \|P_{k,z}\|e^{W_k^\top\tilde{P}_{k,z} + b_k},
           \end{equation}
           where the exponential function is meant ``pointwise''.
        \item Produce the output map~$M_k$ by linear pooling with Gaussian weights:
           \begin{displaymath}
              M_k(z) = \sum_{ u \in \Omega_{\kmone}} e^{-\frac{1}{\beta_k^2} \|u-z\|^2} \tilde{M}_k(u).
           \end{displaymath}
     \end{itemize}
     \State \textsc{Output:} A map $M_k: \Omega_k \to \Real^{p_k}$ . 
  \end{algorithmic}
  \caption{\label{alg:encoding} Encoding layer~$k$ of a CKN.}
  \end{algorithm}

\subsection{\label{sec:sgdopt}Large-Scale Optimization}\label{subsec:ckn6}
One of the challenge we faced to apply CKNs to image retrieval
was the lack of scalability of the original model
introduced by~\citet{mairal2014convolutional}, which was a proof of concept with no effort towards scalability. A first improvement we made was
to simplify the original objective function with changes of variables,
resulting in the formulation~(\ref{eq:cknoptimization}), leading to simpler and less expensive
gradient computations. 

The second improvement is to use stochastic optimization instead of the L-BFGS method used by~\citet{mairal2014convolutional}. This 
allows us to train every layer by using a set of one million patches and
conduct learning on all of their possible pairs, which is a regime where
stochastic optimization is unavoidable.
Unfortunately,
applying stochastic gradient descent directly on~(\ref{eq:cknoptimization})
turned out to be very ineffective due to the poor conditioning of
the optimization problem.
One solution is to make another change of variable and optimize in a space
where the input data is less correlated.

More precisely, we proceed by (i) adding an ``intercept'' (the constant value
1) to the vectors~$x_i$ in~$\Real^q$, yielding vectors~$\tilde{x}_i$ in~$\Real^{q+1}$; (ii)
computing the resulting (uncentered) covariance matrix  $G=\frac{1}{n}\sum_{i=1}^n
\tilde{x}_i\tilde{x}_i^\top$; (iii) computing the eigenvalue decomposition
of~$G=U \Delta U^\top$, where $U$ is orthogonal and $\Delta$ is diagonal with
non-negative eigenvalues; (iv) computing the preconditioning matrix $R=U
(\Delta+\tau I)^{1/2} U^\top$, where $\tau$ is an offset that we choose to be
the mean value of the eigenvalues. 
Then, the matrix $R$ may be used as a preconditioner since the covariance of
the vectors $R\tilde{x}_i$ is close to identity. In fact, it is equal to the
identity matrix when $\tau=0$ and $G$ is invertible.
Then, problem~(\ref{eq:cknoptimization}) with preconditioning becomes
\begin{equation}\label{eq:cknoptimization2}
    \min_{Z} \quad \sum_{i=1}^n \left[e^{\frac{\Vert x_i -x'_i\Vert^2}{2\alpha^2}}
       - \sum_{j=1}^p e^{z_j^\top  R\tilde{x}_i}
    e^{z_j^\top R\tilde{x}'_i} \right] ^2,
 \end{equation}
 obtained with the change of variable
 $[W^\top, b^\top]=Z^\top R$.
 Optimizing with respect to~$Z$ to obtain a solution~$(W,b)$ turned out to be 
the key for fast convergence of the stochastic gradient optimization algorithm.

Note that our effort also consisted on implementing heuristics for 
automatically selecting the learning rate during optimization without 
requiring any manual tuning, following in part standard guidelines from
\citet{bottou2012stochastic}. 
More precisely, we select the initial learning rate in the following range:
$\{1,2^{-1/2},2^{-1},\ldots,2^{-20}\}$,  by performing 1K iterations with 
mini-batches of size $1000$ and choosing the one that gives the lowest
objective, evaluated on a
validation dataset. After choosing the learning rate, we keep
monitoring the objective on a validation set every 1K iteration, and
perform backtracking in case of divergence. The learning rate is also
divided by $\sqrt{2}$ every 50K iterations. 
The total number of iterations is set to 300K.
Regarding initialization, weights are randomly initialized according to a standard normal
distribution.
These heuristics are fixed
over all experiments and resulted in a stable parameter-free learning
procedure, which we will release in an open-source software package.

\subsection{Different Types of CKNs with Different Inputs}\label{sec:inputs} 
We have not discussed yet the initial choice of the map~$M_0$ for representing
an image~$M$. In this paper, we follow~\citet{mairal2014convolutional}
and investigate three possible inputs:
\begin{enumerate}
   \item {\bfseries CKN-raw}:  We use the raw RGB values.  This captures the
      hue information, which is discriminant information for many application
      cases.
   \item {\bfseries CKN-white}: It is similar to CKN-raw with the following
      modification: each time a patch~$P_{0,z}$ is extracted from~$M_0$, it is
      first centered (we remove its mean color), and whitened by computing a PCA on the
      set of patches from $M_0$.  The resulting patches are invariant to the
      mean color of the original patch and mostly respond to local color
      variations.
   \item {\bfseries CKN-grad}: The input~$M_0$ simply carries the
      two-dimensional image gradient computed on graysale values.
      The map has two channels, corresponding to the gradient computed
      along the x-direction and along the y-direction, respectively.
      These gradients are typically computed by finite differences.
\end{enumerate} 
Note that the first layer of CKN-grad typically uses patches of size $1 \times
1$, which are in~$\Real^2$, and which are encoded by the first layer into $p_1$
channels, typically with $p_1$ in $[8;16]$.  This setting corresponds exactly
to the kernel descriptors introduced by~\citet{bo2010kernel}, who have proposed
a simple approximation scheme that does not require any learning. Interestingly,
the resulting representation is akin to SIFT descriptors.

Denoting by $(G_{x,z},G_{y,z})$ the gradient components of image~$M$ at location~$z$, 
the patch $P_z$ is simply the vector $[G_{x,z},G_{y,z}]$ in~$\Real^2$. 
Then, the norm~$\|P_z\|_2$ can be interpreted as the gradient magnitude
$\rho_z = \sqrt{G_{x,z}^2 + G_{y,z}^2}$, and the normalized patch~$\tilde{P}_z$ represents a local orientation.  In fact, there exists~$\theta_z$ such that
$\tilde{P}_z = [\cos(\theta_z), \sin(\theta_z)]$.
Then, we may use the relation~(\ref{eq:rbf}) to approximate the Gaussian kernel
$\exp(-\|\tilde P_z - \tilde P'_{z'}\|^2/(2\alpha_1^2))$.
We may now approximate the integral by sampling $p_1$ evenly 
distributed orientations $\theta_j = 2j\pi/p_1,\,\,:
j\in\{1,\dots,p_1\}$, and we obtain, up to a constant scaling factor,
 \begin{equation}
    e^{-\frac{1}{2\alpha_1^2}\|\tilde P_z - \tilde P'_{z'}\|^2} \approx \sum_{j=1}^{p_1} e^{-\frac{1}{\alpha_1^2}\|\tilde{P}_z - P_{\theta_j}\|^2}e^{-\frac{1}{\alpha_1^2}\|\tilde{P}'_{z'} - P_{\theta_j}\|^2},
 \end{equation}
where $P_\theta=[\cos(\theta),\sin(\theta)]$.
With such an approximation, the $j$-th entry of the map $\tilde{M}_1(z)$
from~(\ref{eq:tmp_map})  should be replaced by
$$ \rho_z  e^{-\frac{1}{2\alpha_1^2}\left((\cos(\theta_j) - \cos(\theta_z))^2+ (\sin(\theta_j) -\sin(\theta_z))^2\right)}.$$
This formulation can be interpreted
as a soft-binning of gradient orientations in a ``histogram'' of size $p_1$ at every location $z$. To
 ensure an adequate distribution in each bin, we choose
$\alpha_1 = \big((1-\cos\left(2\pi/d_1\right))^2 +
\sin\left(2\pi/d_1\right)^2\big)^{1/2}$.
After the pooling stage, the representation becomes very close to SIFT
descriptors.



\begin{figure*}[t!]
\begin{center}
\includegraphics[width=0.9\textwidth]{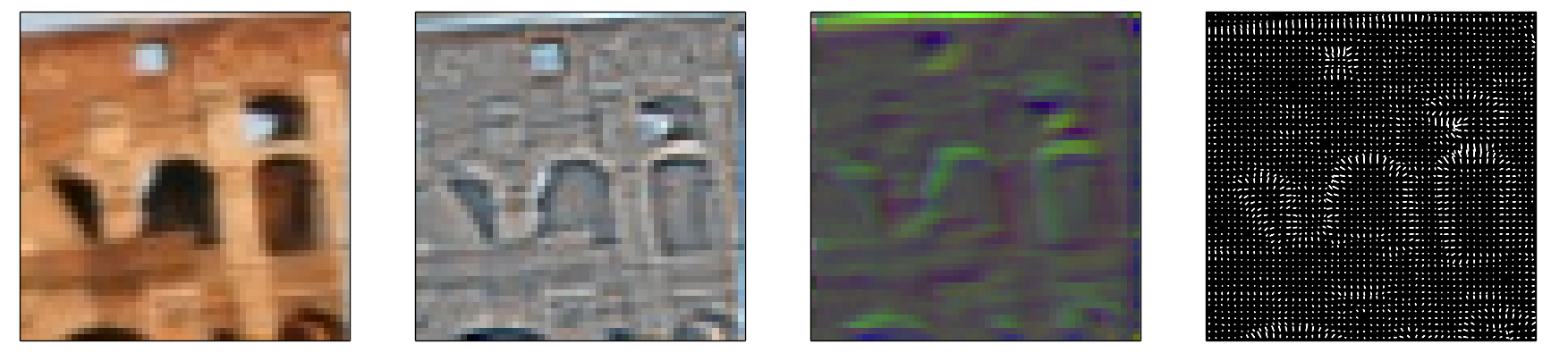}
  \caption{\label{fig:inputs} Vizualization of the possible CKN
    inputs: CKN-raw (left); input when subtracting the mean color of each patch (middle-left); CKN-white, similar to the
    previous but with whitening (middle-right) and CKN-grad
    (right). Patches were extracted with a size of 3x3;
    images are reconstructed by averaging the pixel values of the pre-processed patches
    (up to normalization to fit the 0-256 range). Best viewed in color on a computer screen.} 
\end{center} 
\end{figure*}

A visualization of all input methods 
can be seen in figure~\ref{fig:inputs}. See~\citep{mairal2014sparse}
for more analysis of image preprocessing.

\section{\label{sec:datasets}Datasets} 


In this section, we give details on the standard datasets we use to
evaluate our method, as well as the protocol we used to create our
new dataset.  

\subsection{Standard datasets}

We give details on the commonly used benchmarks for which we report results.

\subsubsection{Patch retrieval}

The dataset introduced in~\citep{mikolajczyk2005comparison} in now standard
to benchmark patch retrieval methods. This dataset consists of a
set of 8 scenes viewed under 6 different conditions, with increasing transformation
strength. In contrast to~\citep{winder2009picking,zagoruyko2015learning} where only DoG patches
are available, the~\citet{mikolajczyk2005comparison} dataset allows custom
detectors.
We extract regions with the Hessian-Affine detector and
match the corresponding descriptors with Euclidean nearest-neighbor. A
pair of ellipses is deemed to match if the projection of the first region
using the ground-truth homography on the second ellipse overlaps by at
least 50\%. The performance is measured in terms of mean average
precision (mAP). 

\subsubsection{Image retrieval}

We selected three standard image retrieval benchmarks: Holidays, Oxford
and the University of Kentucky Benchmark (UKB).

\paragraph{Holidays}

The Holidays dataset~\citep{jegou2008hamming} contains 500 different
scenes or objects, for which 1,491 views are available. 500 images serve as queries. Following common practice, in contrast to~\citep{babenko2014neural} though, we use the unrotated version,
which allows certain views to display a $90^\circ$ rotation with respect to
their query. While this has a non-negligible impact on performance
for dense representations (the authors of~\citep{babenko2014neural} report a 3\% global drop in mAP), this is of
little consequence for our pipeline which uses rotation-invariant keypoints.
External learned parameters, such as
k-means clusters for VLAD and PCA-projection matrix are learned on a
subset of random Flickr images. The standard metric is mAP. 

\paragraph{Oxford}

The Oxford dataset~\citep{philbin2007object} consists of 5,000 images of Oxford landmarks. 11
locations of the city are selected as queries and 5 views per location
is available. The standard benchmarking protocol, which we use, involves
cropping the bounding-box of the region of interest in the query
view, followed by retrieval. Some works, such
as~\citep{babenko2014neural} forgo the last step. 
Such a non-standard protocol yields a boost in performance.
For instance~\cite{babenko2015aggregating} report a close to 6\% improvement with non-cropped queries.
mAP is the standard measure.

\paragraph{UKB} Containing 10,200 photos, the University of Kentucky
Benchmark (UKB)~\citep{nister2006scalable} consists of 4 different
views of the same object, under radical viewpoint changes. All images
are used as queries in turn, and the standard measure is the mean
number of true positives returned in the four first retrieved images ($4\times$recall@$4$).

\subsection{Rome}

One of the goals of this work is to establish a link between
performance in patch retrieval with performance in image
retrieval. Because of the way datasets are constructed differs (e.g.  
Internet-crawled images vs successive shots with the same camera,
range of viewpoint changes, semantic content of the dataset, type of
keypoint detector), patch descriptors may display different
performances on different datasets. We therefore want a dataset
that contains a groundtruth both at patch and image level, to jointly
benchmark the two performances. Inspired by the seminal work
of~\citep{winder2009picking}, we introduce the Rome retrieval dataset,
based on 3D reconstruction of landmarks. The Rome16K
dataset~\citep{li2010location} is a Community-Photo-Collection 
dataset that consists of 16,179 images downloaded from photo sharing
sites, under the search term ``Rome''. Images are partitioned into 66
``bundles'', each one containing a set of viewpoints of a given
location in Rome (e.g. ``Trevi Fountain''). Within a bundle,
consistent parameters were automatically computed and are
available\footnote{\url{www.cs.cornell.edu/projects/p2f}}. The set of
3D points that were reconstructed is also available, but we choose
not to use them in favor of our Hessian-Affine keypoints. To determine
matching points among images of a same bundle, we use the following
procedure. i) we extract Hessian-Affine points in all images. For each
pair of images of a bundle, we match the corresponding SIFTs, using Lowe's
reverse neighbor rule, as well as product
quantization~\citep{jegou2011product} for speed-up. We filter matches,
keeping those that satisfy the epipolar constraint up to a tolerance
of $3$ pixels. Pairwise point matches are then greedily aggregated 
to form larger groups of 2D points viewed from several cameras. Groups
are merged only if the reproduction error from the estimated 3D
position is below the $3$ pixel threshold.

To allow safe parameter tuning, we split the set of bundles into a
train and a test set, respectively containing 44 and 22 bundles. 

\subsubsection{Patch retrieval}

\begin{figure}[t!]
  \begin{center}
    \includegraphics[width = 0.9\linewidth]{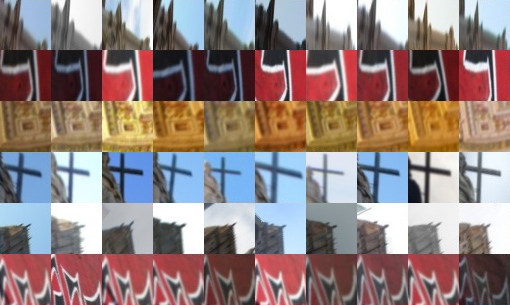}  
    \caption{Examples of patches matched under our procedure. We
      observe significant changes in lighting, but smaller changes in rotation and skew.\label{fig:rome_classes_patch}}

  \end{center}
\end{figure}

We design our patch retrieval datasets by randomly sampling in each train and
test split a set of $1,000$ 3D points for which at least $10$ views
are available. The sampling is uniform in the bundles, which means
that we take roughly the same amount of 3D points from each bundle. We then
sample $10$ views for each point, use one as a query and the remaining
as targets. Both our datasets therefore contain $1,000$ queries and
$9,000$ retrieved elements. We report mean average precision (mAP). An
example of patch retrieval classes can be seen in Fig. \ref{fig:rome_classes_patch}.

\subsubsection{Image retrieval}

\begin{figure*}[t!]
  \begin{center}
    \includegraphics[width= \linewidth]{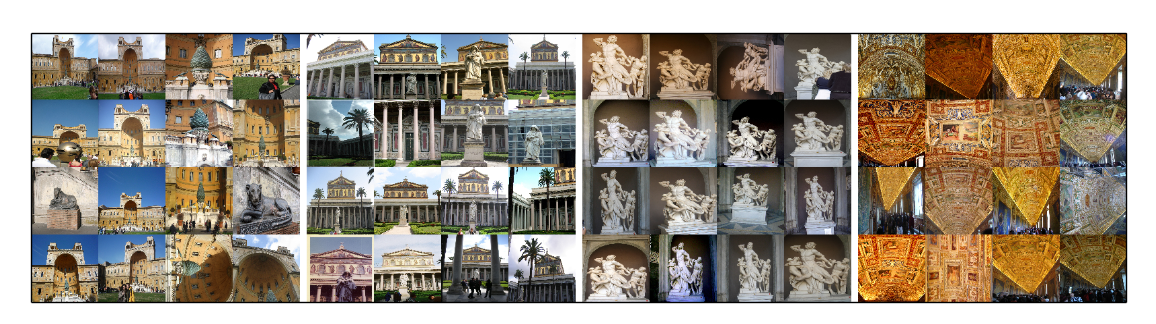}\vspace{-0.5cm}
\caption{\label{fig:rome_classes_image} 
Example of classes of the image retrieval dataset of Rome. Each class consists of a particular location. Some bundle display significant viewpoint changes (extreme left and right), while others have little variation in appearance (middle). Best viewed in color.}
  \end{center}
\end{figure*}

Using the same aforementioned train-test bundle split, we select 1,000
query images and 1,000 target images evenly distributed over all
bundles. Two images are deemed to match if they come from the same
bundle, as illustrated in Fig.~\ref{fig:rome_classes_image}

\section{\label{sec:experiments}Experiments}


In this section, we describe the implementation of our pipelines, and
report results on patch and image retrieval benchmarks.

\subsection{Implementation details}

We provide details on the patch and image retrieval pipelines. Our
goal is to evaluate the performance of patch descriptors, and all
methods are therefore given the same input patches (computed at
Hessian-Affine keypoints), possibly resized to fit the required input
size of the method. We also evaluate all methods with the same aggregation procedure
(VLAD with 256 centroids). We believe that improvements in
feature detection and aggregation are orthogonal to our contribution
and would equally benefit all architectures.

\subsubsection{\label{sec:pipeline}Pipeline}

We briefly review our image retrieval pipeline.

\paragraph{Keypoint detection.}

\begin{figure}[t!]
  \begin{center}
    \includegraphics[width=0.8\linewidth]{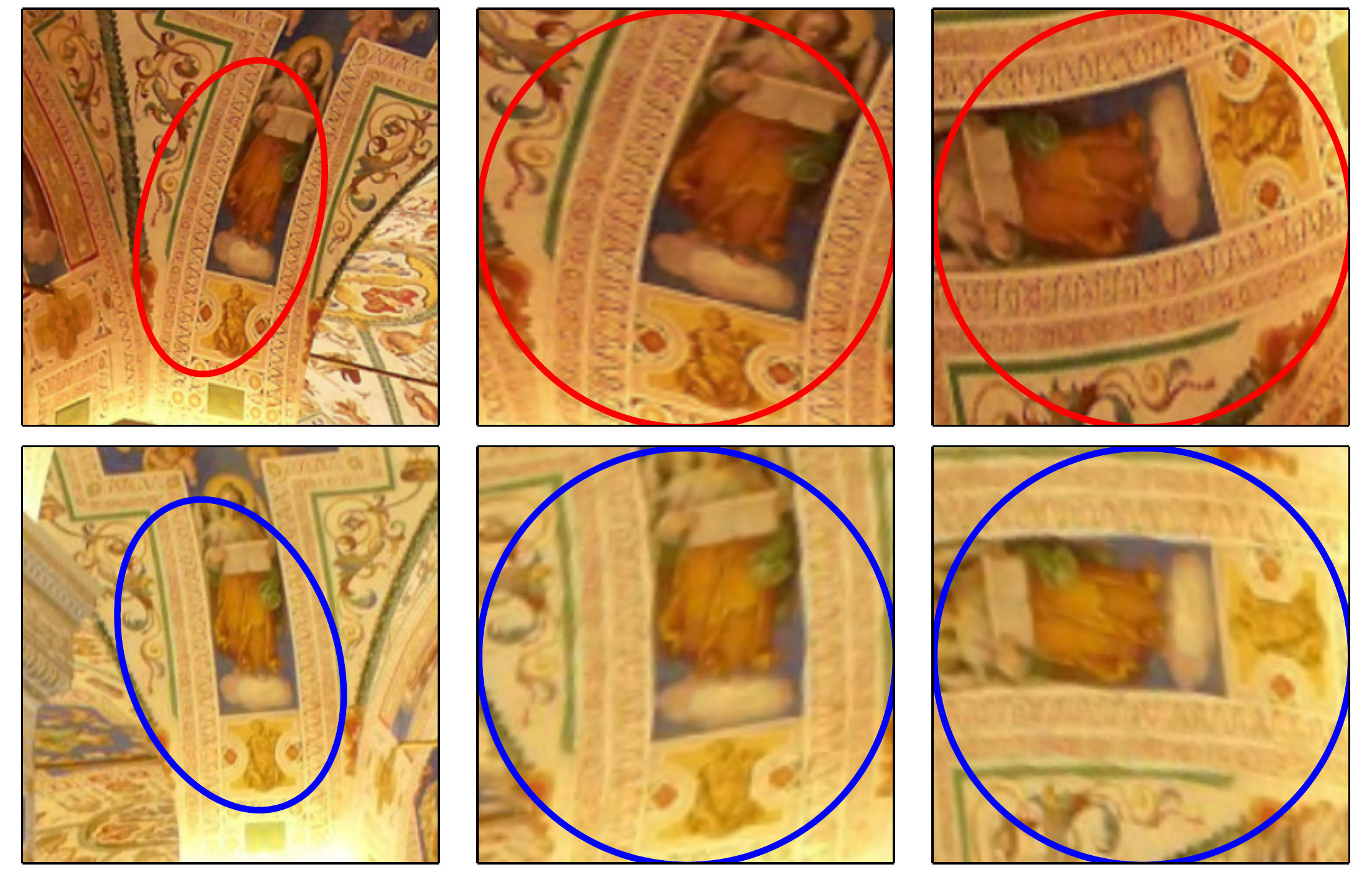}
  \end{center}
\caption{\label{fig:match_affine}Example of matching points in two
  different images. Salient points are extracted (left), affine
  rectified (middle), and normalized in rotation (right). Note that
  the  level of invariance which remains to be covered by the
  patch descriptor is relatively low, as most of the work has been
  accomplished by the detector.}
\end{figure}

A popular design choice in image representations, inspired by text
categorization methods, is to consider images as sets of local
patches, taken at various locations. The choice of 
these locations is left to the interest point detector, for which
multiple alternatives are possible. 

In this work, we use the popular ``Hessian-Affine'' detector of
\cite{mikolajczyk2004scale}. It aims at finding reproducible points,
meaning that selected 3D points of a scene should always belong to the
set of detected image points when the camera undergoes small changes
in settings (e.g. viewpoint, blur, lighting). Because of the ``aperture''
problem the set of such points is limited to textured patches, to the
exclusion of straight lines, for which precise localization is
impossible. This leaves ``blobs'', as used for instance in Lowe's
Difference of Gaussians (DoG) detector~\citep{lowe2004distinctive} or
corners, as is the case for our Hessian-Affine detector. Specifically,
a ``cornerness'' measure is computed on the whole image, based on the
Hessian of the pixel intensities. The set of points whose cornerness is
above a threshold is kept. The detector takes the points at their
characteristic scale and estimates an affine-invariant local
region. Rotation invariance is achieved by ensuring the dominant
gradient orientation always lies in a given direction. This results in
a set of keypoints with locally-affine invariant
regions. Fig.~\ref{fig:match_affine} shows the various steps for
detecting keypoints. We sample the point with a resolution of $51\times 51$ pixels, value that was found optimal for SIFT on Oxford.  Pixels that fall out of the image are set to their nearest
neighbor in the image. This strategy greatly increases patch retrieval
performance compared to setting them to black, as it does not
introduce strong gradients. 

Note that the choice of a particular interest point detector is arbitrary and our method is not specific to Hessian-Affine locations. To show this, we also experiment with dense patches in section~\ref{sec:dense}

\paragraph{Patch description.} Because of the affine-invariant detectors, and as seen in
Fig.~\ref{fig:match_affine}, for a given 3D point seen in two
different images, the resulting patches have small differences
(e.g. lighting, blur, small rotation, skew). The goal of patch
description, and the focus of this work, is to design a patch
representation, {\it i.e.} a mapping $\Phi$ of the space of fixed-size 
patches into some Hilbert space, that is robust to these changes. 

\paragraph{Matching/Aggregation.} Stereo-vision uses this keypoint representation to establish
correspondences between images of the same instance, with 3D
reconstruction as an objective. The cost of this operation is
quadratic in the number of keypoints, which is prohibitive in image
retrieval systems that need to scan through large databases. Instead,
we choose to aggregate the local patch descriptors
into a fixed-length global image descriptor. For
this purpose, we use the popular VLAD
representation~\citep{jegou2012aggregating}. Given a clustering in the
form of a Voronoi diagram with points $\{c_1, \dots, c_k\}$ of the feature space (typically obtained using k-means
on an external set of points), VLAD encodes the set of visual words
$\{x_1,\dots,x_n\}$ as the total shift with respect to their assigned
centroid: 
\begin{equation}
  {\rm VLAD} = \sum_{i=1}^k\sum_{j=1}^n \varepsilon_{i,j}(x_j - c_i),
\end{equation}

where $\varepsilon_{i,j}$ is the assignment operator, which is 1 if
$x_j$ is closer to centroid $c_i$ than to the others, and 0
otherwise.

The final VLAD descriptors is
power-normalized with exponent 0.5 (signed square-root), as well as
$\ell_2$-normalized.

\subsubsection{Convolutional Networks training}
\paragraph{AlexNet-ImageNet.} We use the Caffe
package~\citep{jia2014caffe} and its provided weights for AlexNet, that
have been learned according
to~\cite{krizhevsky2012imagenet}. Specifically, the network was
trained on ILSVRC'12 data for 90 epochs, using a learning
rate initialized to $10^{-2}$ and decreased three times prior to
termination. Weight decay was fixed to 0.0005, momentum to 0.9 and
dropout rate to 50\%. It uses three types of image jittering: random
$227 \times 227$ out of $256 \times 256$ cropping, flip and color variation.  

\paragraph{AlexNet-Landmarks.} Following~\cite{babenko2014neural}, we
fine-tune AlexNet using images of the Landmarks
dataset\footnote{\url{http://sites.skoltech.ru/compvision/projects/neuralcodes/}}. Following
their protocol, we strip the network of its last layer and replace it
with a 671-dim fully-connected one, initialized with random Gaussian
noise (standard deviation 0.01). Other layers are initialized with the old AlexNet-ImageNet weights. 
We use a learning rate of $10^{-3}$ for fine-tuning. 
Weight decay, momentum and dropout rate are kept to their default values (0.0005, 0.9 and 0.5).  
We use data augmentation at training time, with the same
transformations as in the original paper (crop, flip and color). We
decrease the learning rate by 10 when it saturates (around 20 epochs each
time). We report a validation accuracy of 59\% on the
Landmarks dataset, which was confirmed through discussion with the
authors. On Holidays, we report a mAP of 77.5 for the sixth
layer (against 79.3 in~\citep{babenko2014neural}), and 53.6 (against
54.5) on Oxford. Even though slightly below the results in the
original paper, fine-tuning still significantly improves on ImageNet
weights for retrieval.

\begin{figure*}[t]
  \begin{center}
    \includegraphics[angle=90,width=0.7\linewidth]{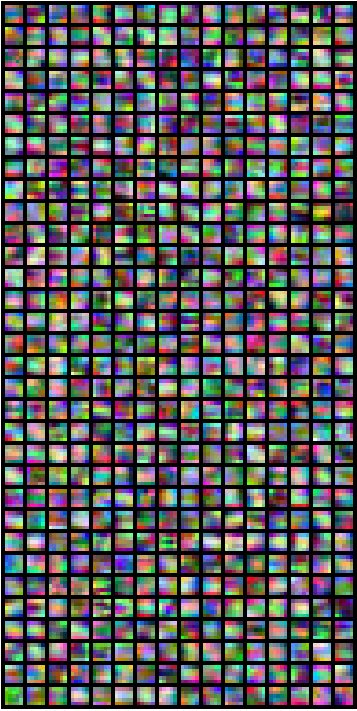}
    \caption{\label{fig:ckn_raw_kernels} All 512 kernels of the first layer of our CKN-raw architecture.}
  \end{center}
\end{figure*}

\paragraph{PhilippNet.} For PhilippNet, we used the model provided by
the authors. The model is learned on 16K surrogate classes (randomly
extracted patches) with 150 representatives (composite
transformations, including crops, color and contrast variation, blur,
flip, etc.). We were able to replicate their patch retrieval results on
their dataset, as well as on \mikodatspace when using MSER
keypoints. 
\paragraph{PhilippNet-Rome.} The patch retrieval dataset of RomePatches
does not contain enough patches to learn a deep network. We augment it
using patches extracted in a similar fashion, grouped in classes that
correspond to 3D locations and contain at least 10 examples. We build
two such training sets, one with 10K classes, and one with 100K
classes. Training is conducted with the default parameters. 
\paragraph{Deepcompare.} As previously described, we use the online
code provided by~\cite{zagoruyko2015learning}. It
consists of networks trained on the three distinct datasets
of~\cite{winder2009picking}: Liberty, NotreDame and Yosemity. For our
image retrieval experiments, we can only use the siamese networks, as
the others do not provide a patch representation. These were observed
in the original paper to give suboptimal results.  
\paragraph{Convolutional Kernel Networks} To train the convolutional kernel networks, 
we randomly subsample a set of 100K
patches in the train split of the Rome dataset. For each layer, we
further extract 1M sub-patches with the required size, and feed all
possible pairs as input to the CKN. The stochastic gradient
optimization is run for 300K iterations with a batch size of
1000, following the procedure described in section~\ref{sec:sgdopt}.
Training a convolutional kernel network, for a particular set of hyperparameters, 
roughly takes 10 min on a GPU. This stands in contrast to the 2-3 days for 
training using the L-BFGS implementation used in~\cite{mairal2014convolutional}.
We show in Fig.~\ref{fig:ckn_raw_kernels} a visualization of the first
convolutional filters of CKN-raw. The sub-patches for CKN-grad and
CKN-white are too small ($3 \times 3$) and not adapted to viewing.  

\subsection{Patch retrieval}

\subsubsection{CKN parametric exploration}

The relatively low training time of CKNs (10 min. on a recent
GPU), as well as the fact that the training is layer-wise -- and therefore
lower layer parameters can be reused if only the top layer change -- 
allows us to test a relatively high number of parameters, and select
the ones that best suit our task. We tested parameters for
convolution and pooling patch sizes in range of $2$ to $5$, number of
neuron features in powers of $2$ from 128 to 1024. For the $\sigma$
parameter, the optimal value was found to be $10^{-3}$ for all
architectures. For the other parameters, we keep one set of optimal
values for each input type, described in Table~\ref{tab:CKNarch}.

\begin{table}[ht] 
\begin{center}
  \begin{tabular}{|l|c|c|c|}\hline            
    \bf Input & \bf Layer 1 & \bf Layer 2 & dim. \\\hline
    \bf CKN-raw & 5x5, 5, 512 & ---- &  41,472\\
      \bf CKN-white & 3x3, 3, 128 & 2x2, 2, 512 & 32,768 \\
    \bf CKN-grad &  1x1, 3, 16 & 4x4,2,1024 & 50,176 \\\hline
  \end{tabular}
\end{center}
\caption{\label{tab:CKNarch} 
  For each layer we indicate the sub-patch size, the subsampling factor 
  and the number of filters. For the gradient network, the value $16$ corresponds to the number of orientations.}
\end{table}

\begin{figure}[t]
  \begin{center}
    \includegraphics[width=0.9\linewidth]{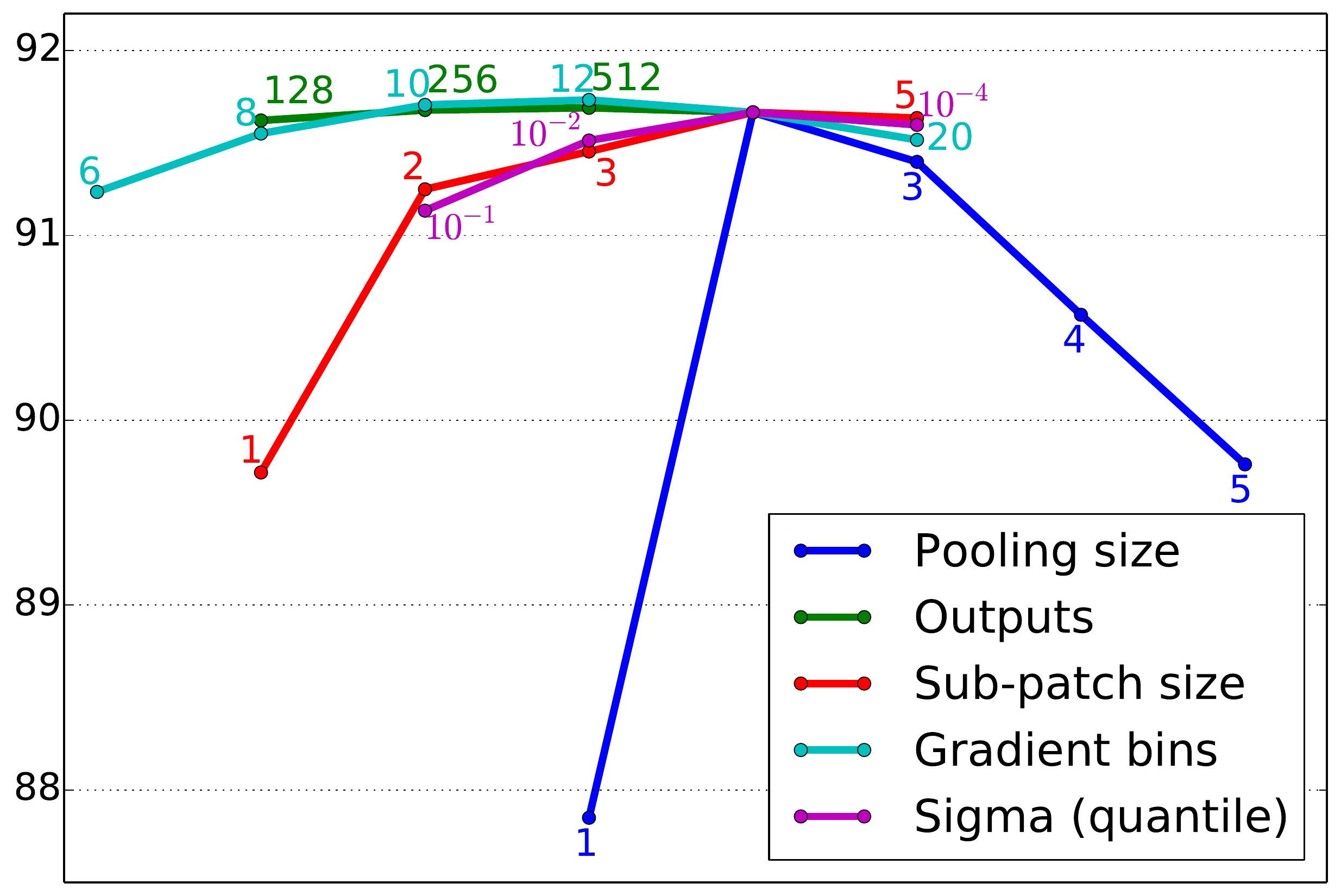}
\caption{\label{fig:exploration} mAP results on the train set of
  RomePatches with CKN-grad, whose hyper-parameters have been changed
  one by one around the optimal point. Note that 10 and 12 gradient bins give
  slightly better results but 16 was kept to align with the logarithm scale of the grid.}
  \end{center}
\end{figure}

In Figure~\ref{fig:exploration}, we explore the impact of the
various hyper-parameters of CKN-grad, by tuning one while keeping the
others to their optimal values. 

\subsubsection{Dimensionality reduction}

Since the final dimension of the CKN descriptor is prohibitive for most applications of practical value
(see Table \ref{tab:CKNarch}), we investigate dimensionality reduction techniques. 
Note that this step is unnecessary for CNNs, whose feature dimension do not
exceed 512 (PhilippNet). We only perform unsupervised dimensionality
reduction through PCA, and investigate several forms of whitening.
Denoting $X$ the $n\times d$ matrix of $n$ CKN features (we used
$n=10K$), the singular value decomposition of $X$ writes as 
\begin{equation}
  X = U S V^\top,
\end{equation}

where $U$ and $V$ are orthogonal matrices and $S$ is diagonal with
non-increasing values from upper left to bottom right. For a new
$n'\times d$ matrix of observations $X'$, the dimensionality
reduction step writes as
\begin{equation}
  X'_{\rm proj} = X L^\top,
\end{equation}
where $L$ is a $d'\times d$ projection matrix. 

The three types of whitening we use are: i) no whitening; ii) full whitening; iii) semi-whitening. 
Full whitening corresponds to 
\begin{equation}
  L = V(1:d',:) / D(1:d',1:d') \; ,
\end{equation}
while semi-whitening corresponds to 
\begin{equation}
  L = V(1:d',:) / \sqrt{D(1:d',1:d')} \; .
\end{equation}

The matrix division $/$ denotes the matrix multiplication with the
inverse of the right-hand matrix. The square-root is performed element-wise.

\begin{figure}
\newcommand{\igpca}[1]{\hspace*{-3mm}\includegraphics[width=3.3cm]{CKN-#1.pdf}}
\begin{center}
\hspace*{-5mm}
\begin{tabular}{c@{}c@{}c}
CKN-grad & CKN-raw & CKN-white \\
\igpca{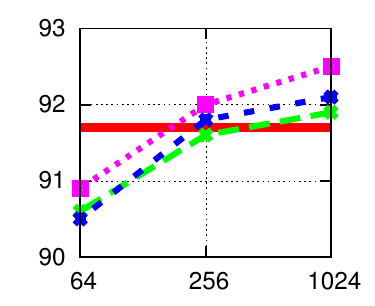}  & 
\igpca{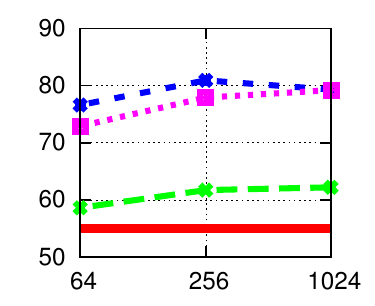}  & 
\igpca{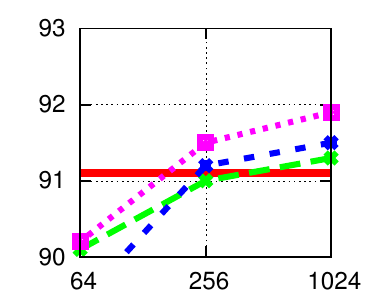}  \\
\end{tabular}
\includegraphics[width=6cm]{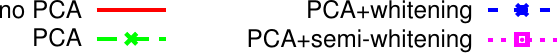}
\end{center}
\caption{\label{fig:CKNPCA} Influence of dimensionality reduction 
  on patch retrieval performance. Results reported in mAP (\%) on the train split of
  RomePatches as a function of the PCA dimension. As a comparison, SIFT reports 91.6\%.}   
\end{figure}

Results of the different methods on the RomePatches dataset are displayed in
Fig.~\ref{fig:CKNPCA}. We observe that semi-whitening works best for
CKN-grad and CKN-white, while CKN-raw is slightly improved by full
whitening. We keep these methods in the remainder of this article, as
well as a final dimension of 1024.

\subsubsection{Results}

We compare the convolutional architectures on our three patch
datasets: RomePatches-train, RomePatches-test and \mikodat. Results
are given in Table \ref{tab:patch_results}. For AlexNet CNNs, we
report results for all outputs of the 5 convolutional layers (after
ReLU). We note that SIFT is an excellent baseline for these
methods, and that CNN architectures that were designed for local
invariances perform better than the ones used in AlexNet, as observed
by \cite{fischer2014descriptor}. The results of the PhilippNet on \mikodatspace are 
different from the ones reported by~\cite{fischer2014descriptor},
as we evaluate on Hessian-Affine
descriptors while they use MSER. To have a comparable setting,
we use their network with an input of 64x64 that corresponds to the
coverage of one top neuron, as well as their protocol that slide it on
91x91 patches. We notice that this last step only provides a small
increase of performance (2\% for patch retrieval and 1\% for image
retrieval). We observe that PhilippNet outperforms both 
SIFT and AlexNet, which was the conclusion of~\cite{fischer2014descriptor}; CKN
trained on whitened patches do however yield better results.
 
\begin{table}[t!] 
\resizebox{\linewidth}{!}{%
\begin{tabular}{|l|cc|c|c|c|}\hline
  \bf Architecture & \bf coverage & \bf  Dim & \multicolumn{2}{c|}{\bf RomePatches} &\bf  Miko. \\
                 &  &          &  \bf train & \bf test           &                 \\\hline
  SIFT & 51x51 & 128 & 91.6 & 87.9  & 57.8 \\\hline
  AlexNet-conv1 & 11x11 & 96 & 66.4 & 65.0 & 40.9 \\
  AlexNet-conv2 & 51x51 & 256 & 73.8 & 69.9 & 46.4 \\
  AlexNet-conv3 & 99x99 & 384 & 81.6 & 79.2 & 53.7 \\
  AlexNet-conv4 & 131x131 & 384 & 78.4 & 75.7 & 43.4 \\
  AlexNet-conv5 & 163x163 & 256 & 53.9 & 49.6 & 24.4\\\hline
  PhilippNet & 64x64 & 512 & 86.1 & 81.4 & 59.7 \\
  PhilippNet & 91x91 & 2048 & 88.0 & 83.7 & 61.3 \\\hline
  CKN-grad & 51x51 & 1024 & \bf 92.5 & \bf 88.1 & 59.5\\
  CKN-raw & 51x51 & 1024 & 79.3 & 76.3 & 50.9\\
  CKN-white & 51x51 & 1024 & 91.9 & 87.7 & \bf 62.5\\\hline
\end{tabular}  
}
\caption{\label{tab:patch_results}Results of convolutional
  architectures for patch retrieval.} 
\end{table}

\subsubsection{DeepCompare} 

As the architectures of
DeepCompare~\citep{zagoruyko2015learning} do not rely on an underlying
patch descriptor representation but rather on similarities between
pairs of patches, some architectures can only be tested for patch-retrieval. 
We give in Table \ref{tab:deepcompare} the performances of all
their architectures on RomePatches. 

\begin{table}[ht!]
\begin{center}
  \resizebox{\linewidth}{!}{\begin{tabular}{|c|c|c|}\hline
    \bf Network & \bf RomePatches-Train& \bf RomePatches-Test\\\hline
    2ch2stream Y&	89.3&	85.7\\
    2ch2stream ND&	88.3&	84.9\\
    2ch2stream L&	\bf 90.2&\bf 86.6\\\hline
    2ch Y&	86.6&	83.5\\
    2ch ND&	81.2&	78.4\\
    2ch L&	83.4&	80.1\\\hline
    siam Y& 82.9 & 79.2 \\
    siam ND& 84.8 & 81.0\\
    siam L& 83.2 & 79.8 \\\hline
    siam2stream Y&	78.8&	75.4\\
    siam2stream ND&	84.2&	80.6\\
    siam2stream L&	82.0&	78.6\\\hline
  \end{tabular}}

\caption{\label{tab:deepcompare}Evaluation of the
    deepcompare architectures on RomePatches in mAP (\%). Networks were trained on
  the three subsets of the Multi-view Stereo Correspondence Dataset:
  Yosemity (Y), Notre-Dame (ND) and Statue of Liberty (L). The
  network notations are as in the original
  paper~\citep{zagoruyko2015learning}.}
\end{center}
\end{table}

We note that the only networks that produce descriptors that can be
used for image retrieval are the architectures denoted ``siam''  here. They also
perform quite poorly compared to the others. Even the best
architectures are still below the SIFT baseline (91.6/87.9).

The method of~\cite{zagoruyko2015learning} optimizes and tests on
different patches (DoG points, 64x64 patches, greyscale), which
explains the poor performances when transferring them to our
RomePatches dataset. The common evaluation protocol on this dataset
is to sample the same number of matching and non-matching pairs, rank
them according to the measure and report the false positive
rate at 95\% recall (``FPR@95\%'', the lower the better). The patches used for the
benchmark are available online, but not the actual split used for
testing. With our own split of the Liberty dataset, we get the
results in Table~\ref{tab:daisypatches}. We note that all our results,
although differing slightly from the ones
of~\cite{zagoruyko2015learning}, do not change their conclusions. 

\begin{table}[t]
\begin{center}
  \begin{tabular}{|c|c|}\hline
    \bf Descriptor & \bf FPR@95\% (\%)\\\hline
    SIFT & 19.9 \\
    Best AlexNet (3) & 13.5\\\hline
    siam-l2 (trained on Notre-Dame)&14.7\\
    2ch-2stream (trained on Notre-Dame)& \bf 1.24 \\\hline
    CKN-grad & 27.7 \\
    CKN-white & 30.4 \\
    CKN-raw & 41.7 \\\hline
    best CKN & 14.4 \\\hline
  \end{tabular}
\caption{\label{tab:daisypatches} Experiments on a set of pairs of the
  Liberty dataset. Measure is false positive rate @95\% recall, and
  therefore the lower the better. The third layer of AlexNet is used,
  as it provides the best results.}
\end{center}
\end{table}

We note that our CKNs whose hyperparameters were optimized on RomePatches
work poorly. However, optimizing again the parameters for CKN-grad on
a grid leads to a result of 14.4\% (``best CKN''). To obtain this
result, the number of gradient histogram bins was reduced to 6, and
the pooling patch size was increased to 8 on the first layer and 3 on the
second. These changes indicate that the DoG keypoints are less
invariant to small deformations, and therefore require less rigid
descriptors (more pooling, less histograms). The results are on par
with all comparable architectures (siam-l2: 14.7\%, AlexNet: 13.5\%),
as the other architectures either use central-surround networks or
methods that do not produce patch representations. The best method
of~\cite{zagoruyko2015learning}, 2ch-2stream, reports an impressive
1.24\% on our split. 
However, as already mentioned, this architecture does not produce a patch representation
and it is therefore difficult to scale to large-scale patch and image retrieval applications.

\subsubsection{Impact of supervision} We study the impact of the
different supervised trainings, between surrogate classes and real
ones. Results are given in Table \ref{tab:supervision}.

\begin{table}[ht!]
  \begin{center}
    \resizebox{\linewidth}{!}{\begin{tabular}{|c|c|c|}\hline
      \bf Training parameters& \bf RomePatches-Train &\bf RomePatches-Test \\\hline
      PhilippNet&	86.1&	81.4\\
      PhilippNet-Rome 10K&	84.1&	80.1\\
      PhilippNet-Rome 100K&	89.9&	85.6\\\hline
      (Best) CKN-grad&	92.5&	88.1\\
      (Baseline) SIFT&	91.6&	87.9\\\hline
    \end{tabular}}

\caption{Impact of supervision on patch retrieval. PhilippNet is
  trained on surrogate classes, while PhilippNet-Rome is trained on a
  larger set of RomePatches-Train, containing either 10K or 100K
  classes. We see that retraining improves performance
  provided enough data is given. Supervised CNNs
  are still below the SIFT baseline, as well as the unsupervised
  CKNs.\label{tab:supervision}}

  \end{center}
\end{table}

\begin{figure}[t!]
  \begin{center}
    \includegraphics[width=\linewidth]{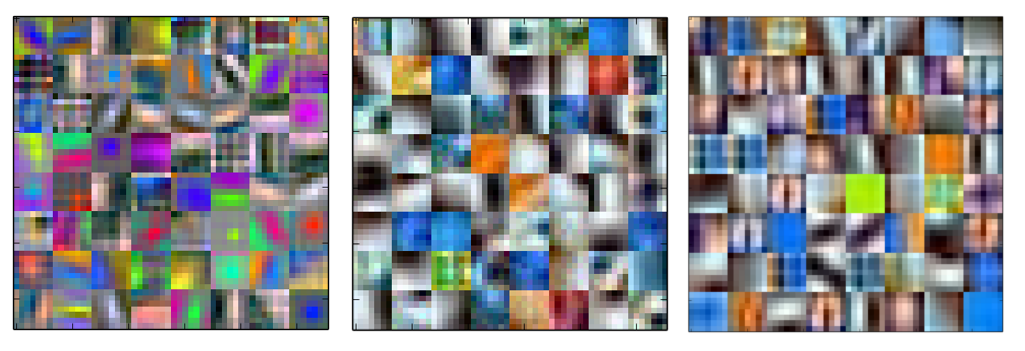}
  \end{center}
\caption{\label{fig:brox_filters} First convolutional filters of the
  PhilippNet learned with surrogate classes (left), 10K Rome classes
  (middle) and 100K Rome classes. Best viewed in color.}
\end{figure}

Figure \ref{fig:brox_filters} gives the first convolutional filters of
the PhilippNet learned on surrogate classes, and on Rome. As can be
seen, the plain surrogate version reacts to more diverse color, as it
has been learned with diverse (ImageNet) input images. Both Rome 10K
and Rome 100K datasets have colors that correspond to skies and
buildings and focus more on gradients. It is interesting to note that
the network trained with 100K classes seems to have captured finer
levels of detail than its 10K counterpart. 

\subsubsection{Robustness}

\begin{figure*}
  \begin{center}
    \includegraphics[width=0.32\linewidth]{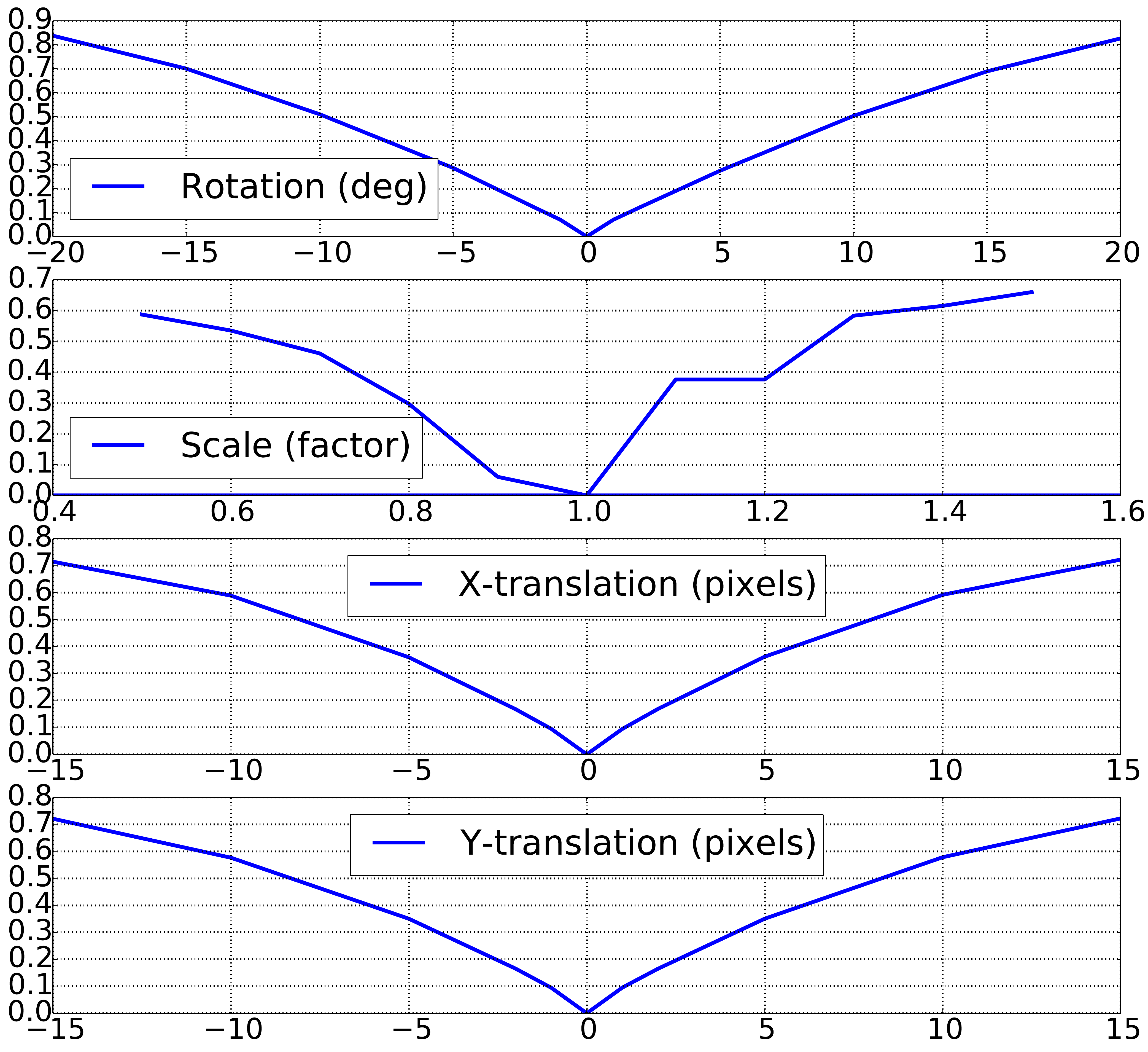}
    \includegraphics[width=0.32\linewidth]{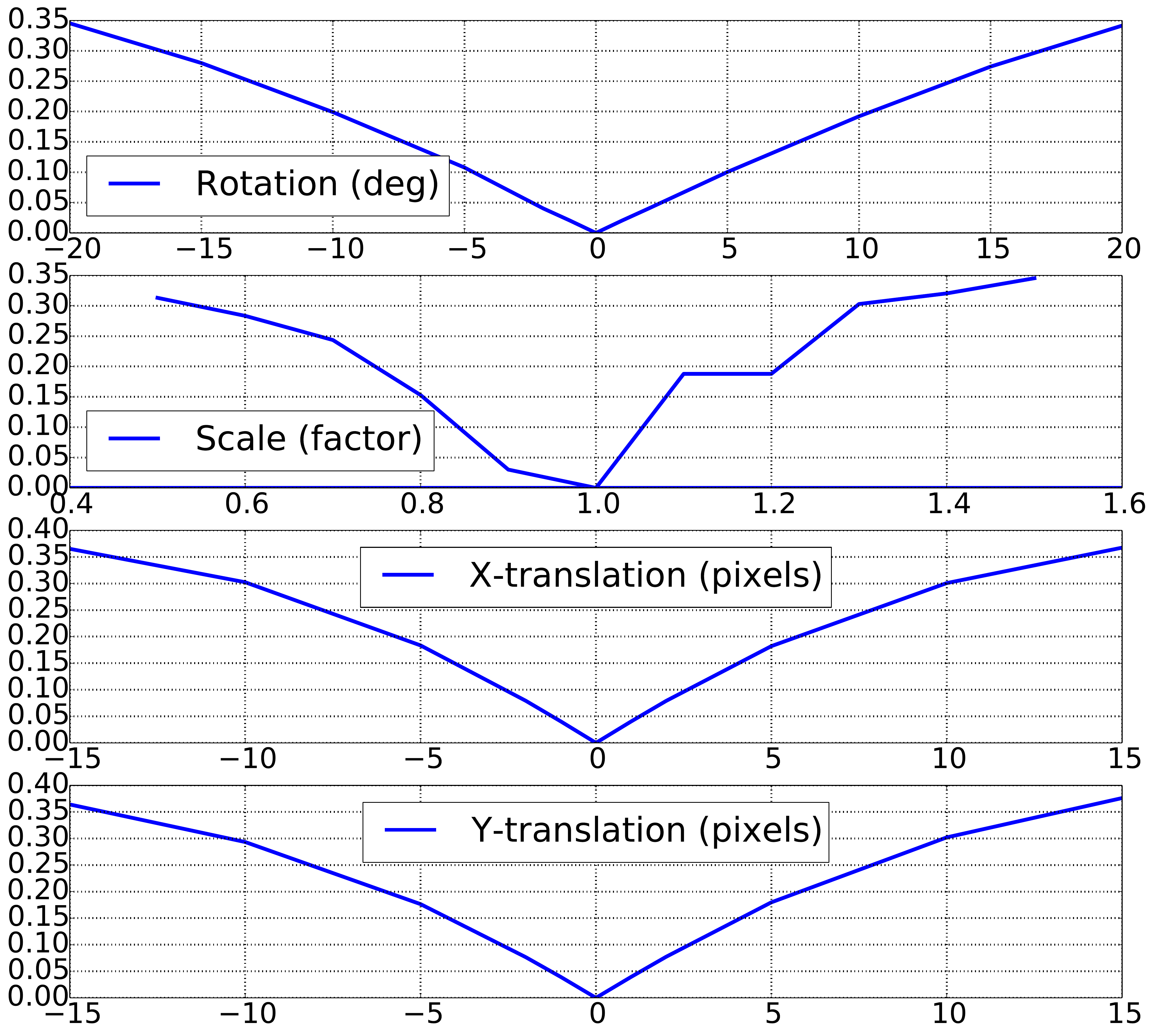}
    \includegraphics[width=0.32\linewidth]{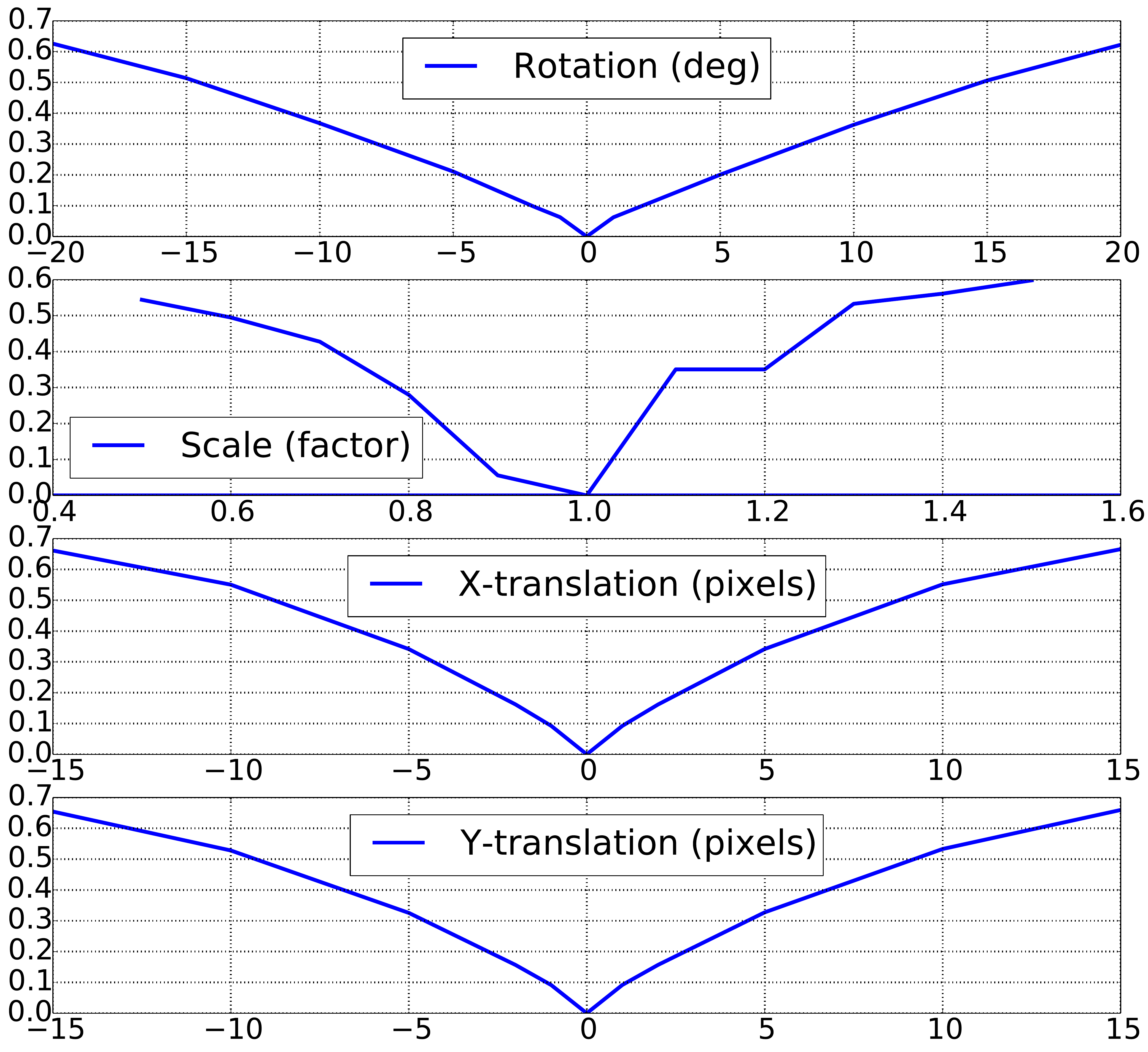}
\caption{\label{fig:robustness} Impact of various transformations on
  CKN descriptors. Represented is the average distance between a patch descriptor
  and its jittered version, as a function of the transformation
  magnitude. Best viewed in numerical format.}
  \end{center}
\end{figure*}

We investigate in Fig.~\ref{fig:robustness} the robustness of CKN
descriptors to transformations of the input patches, specifically rotations, zooms and
translations. We select 100 different images, and
extract the same patch by jittering the keypoint along the
aforementioned transformation. We then plot the average $L_2$ distance
between the descriptor of the original patch and the transformed one. 
Note the steps for the scale transformation; this is due to the fact
that the keypoint scale is quantized on a scale of powers of 1.2, for
performance reasons. 

\subsection{Image retrieval}

\subsubsection{Settings.} 

We learn a vocabulary of $256$
centroids on a related database: for Holidays and UKB we use 5,000
Flickr images and for Oxford, we train on
Paris~\citep{philbin2008lost}. The vocabulary for RomePatches-Train is
learned on RomePatches-Test and vice-versa. The final VLAD descriptor
size is 256 times the local descriptor dimension. 

\subsubsection{Results}

We compare all convolutional approaches as well as the SIFT baseline
in the image retrieval settings. Results are summarized in Table
\ref{tab:imresults}. 
 
\begin{table}[t!]
\begin{center}
  \resizebox{\linewidth}{!}{ 
    \begin{tabular}{|l|c|c|c|c|c|}\hline
      & \bf Holidays &\bf UKB &\bf Oxford &\multicolumn{2}{c|}{\bf Rome}\\
      & & & & train & test \\ \hline
      
      \bf SIFT & 64.0 & 3.44 & 43.7 & 52.9 & 62.7 \\\hline
      \bf AlexNet-conv1 & 59.0 & 3.33 & 18.8 & 28.9 & 36.8 \\   
      \bf AlexNet-conv2 & 62.7 & 3.19 & 12.5 & 36.1 & 21.0\\
      \bf AlexNet-conv3 & \bf 79.3 & 3.74 & 33.3 & 47.1 & 54.7 \\
      \bf AlexNet-conv4 & 77.1 & 3.73 & 34.3 & 47.9 & 55.4 \\
      \bf AlexNet-conv5 & 75.3 & 3.69 & 33.4 & 45.7 & 53.1 \\\hline  
      \bf PhilippNet 64x64 & 74.1 & 3.66 & 38.3 & 50.2 & 60.4\\
      \bf PhilippNet 91x91 & 74.7 & 3.67 & 43.6 & 51.4 & 61.3\\\hline 
      \bf CKN-grad  & 66.5 & 3.42 & \bf 49.8 & \bf 57.0 & \bf 66.2\\
      \bf CKN-raw & 69.9 & 3.54 & 23.0 & 33.0 & 43.8 \\ 
      \bf CKN-white &78.7 & 3.74 & 41.8 & 51.9 & 62.4 \\\hline 
      \bf CKN-mix & \bf 79.3 & \bf 3.76 & 43.4 & 54.5 & 65.3 \\\hline  
          
  \end{tabular}}
\end{center}
\caption{\label{tab:imresults}Image retrieval results. 
Results are in mAP except for UKB where we measure the average
number of true positives in the first 4 results.  
CKN-mix is the result of the concatenation of the VLAD descriptors for
the three channels.} 
\end{table}

On datasets for which color is dominant (e.g. Holidays or UKB), the best
individual CKN results are attained by CKN-white, improved by 
combining the three channels. 
On images of buildings, gradients still perform best
and the addition of color channels is harmful, which explains on the one hand the
poor performance of AlexNet and on the other hand the relatively good performance of PhilippNet 
which was explicitly trained to be invariant to colorimetric transformations.

\subsubsection{Influence of context}

Through experiments with AlexNet-landmarks, we study the impact of
context in the local features. Specifically, we test whether
fine-tuning on a dataset that shares semantic information with the
target retrieval dataset improves performance. 
We compare the same network architecture -- AlexNet -- in two settings:
when parameters are learned on ImageNet (which involves a varied set of classes) 
and when they are learned on the Landmarks dataset which solely consists of buildings and places.
Results, shown in Table~\ref{tab:landmarks}, show clear improvement
for Oxford, but not for Holidays. 
We explain this behavior by the fact that the network learns building-specific invariances 
and that fewer building structures are present in Holidays as opposed to Oxford.

\begin{table}[t!]
\resizebox{\linewidth}{!}{%
  \begin{tabular}{c|c|c||c|c|}\cline{2-5}
    \multirow{2}{*}{} & \multicolumn{2}{c||}{\bf Holidays}&\multicolumn{2}{c|}{\bf Oxford}\\\cline{2-5} 
    & ImageNet & Landmarks & ImageNet & Landmarks\\\hline
\multicolumn{1}{|c|}{conv1} & 59.0 & 57.7 & 18.8 & 20.7 \\ 
\multicolumn{1}{|c|}{conv2} & 62.7 & 72.4 & 12.5 & 23.8 \\
\multicolumn{1}{|c|}{conv3} & 79.3 & 76.1 & 33.3 & 38.6 \\
\multicolumn{1}{|c|}{conv4} & 77.1 & 73.9 & 34.3 & 38.3 \\
\multicolumn{1}{|c|}{conv5} & 75.3 & 66.1 & 33.4 & 32.5 \\\hline
  \end{tabular}}
\caption{\label{tab:landmarks} Influence of the pretraining dataset on
  image retrieval. With the same architecture (AlexNet), training is
  conducted either on ILSVRC'12 data or on the Landmarks
  dataset. Using semantic information related to buildings and places yields 
  improvements for Oxford, but not for Holidays.}
\end{table}

\subsubsection{Dimensionality reduction}

As observed in previous work~\citep{jegou2012whiten}, projecting
 the final VLAD descriptor to a lower dimension using PCA+whitening can
 lead to lower memory costs, and sometimes to slight increases in
 performance (e.g.\ on Holidays~\citep{gong2014multi}). We project to 4096-dim descriptors,
 the same output dimension as ~\cite{babenko2014neural} and obtain
 the results in Table~\ref{tab:pca}.
We indeed observe a small improvement on Holidays but a small decrease on Oxford.

 \begin{table}[t]
   \begin{center}
    \scalebox{0.95}{\begin{tabular}{|c|c|c|c|}\hline
       \bf Dataset & \bf Holidays& \bf UKB &
                                                                      \bf Oxford \\
       &  (CKN-mix)  & (CKN-mix)  & (CKN-grad) \\ \cline{2-4}
       Full (262K-dim)& 79.3 & 3.76 & 49.8 \\
       PCA (4096-dim)& 82.9 & 3.77 & 47.2 \\\hline
     \end{tabular}}
   \end{center}
     \caption{\label{tab:pca} Impact of dimensionality reduction by
       PCA+whitening on the best channel for each dataset. PCA to 4096
       dimensions.}

 \end{table}

\subsubsection{\label{sec:dense}Dense keypoints}

Our choice of Hessian-Affine keypoints is arbitrary and can be
suboptimal for some image retrieval datasets. Indeed we observe that
by sampling points on a regular grid of $8 \times 8$ pixels at multiple
scales, we can improve results. We learn the CKN parameters
and the PCA projection matrix on a dense set of points in Rome, and
apply it to image retrieval as before. We use SIFT descriptors as a baseline,
extracted at the exact same locations. We observe that for CKN-grad
and CKN-white, the previous models lead to suboptimal
results. However, increasing the pooling from 3 (resp. 2 for the
second layer) to 4 (resp. 3), leads to superior performances. We
attribute this to the fact that descriptors on dense points require more
invariance than the ones computed at Hessian-Affine locations. 
Results on the Holidays dataset are given in Table~\ref{tab:dense}.
As observed on Hessian-Affine points, the gradient channel performs
much better on Oxford. We therefore only evaluate this channel. As
explained in section~\ref{sec:datasets}, there
are two ways to evaluate the Oxford dataset. The first one crops
queries, the second does not. While we only consider the first
protocol to be valid, it is interesting to investigate the second, as
results in Table~\ref{tab:oxford_dense} tend to indicate that it
favors dense keypoints (and therefore the global descriptors of~\cite{babenko2014neural}).

\begin{table}[t]
  \resizebox{\linewidth}{!}{%
    \begin{tabular}{|c|c||c|c|c|c|}\hline
    \bf Descriptor & SIFT & grad & raw & white & mix \\\hline
    Hessian-Affine & 64.0 & 66.5 & 69.9 & 78.7 & 79.3 \\
    Dense (same parameters) & 70.3 & 68.5 & 72.3 & 76.8 & .\\
    Dense (changed pooling) & 70.3 & 71.3 & 72.3 & 80.8 & 82.6 \\\hline
  \end{tabular}}
\caption{\label{tab:dense} Dense results on Holidays. Right hand side
  of the table are CKN descriptors. ``Same parameters'' correspond to CKNs
  that have the same parameters as Hessian-Affine ones, yet learned on
  densely extracted patches. ``Changed pooling'' have pooling size
  increased by one at each layer (CKN-raw is unchanged as it only has
  one layer, and already gives good performances).} 
\end{table}

\begin{table}[t]
\begin{center}
  \begin{tabular}{|c|c|c|}\hline
   \bf Method $\backslash$ descriptor & SIFT & CKN-grad \\\hline
    Hessian-Affine, no crop & 45.7 & 49.0 \\
    Hessian-Affine, crop & 43.7 & 49.8 \\\hline
    Dense, no crop & 51.1& 55.4\\
    Dense, crop& 47.6 & 50.9 \\\hline
  \end{tabular}
\end{center}
\caption{\label{tab:oxford_dense} Dense keypoints on the Oxford
  dataset. ``Crop'' indicates the protocol where queries are cropped
  to a small bounding-box containing the relevant object, while ``no
  crop'' takes the full image as a query. For CKN, parameters have
  increased pooling sizes for dense keypoints, as on Holidays.}
\end{table}

\subsubsection{Comparison with state of the art}

Table~\ref{tab:imresultsSOA} compares our approach to recently published results.
Approaches based on VLAD with SIFT \citep{arandjelovic2013all,
jegou2012aggregating}
can be improved significantly  by CKN local
descriptors (+15\% on Holidays). 
To compare to the state of the art with SIFT on
Oxford~\citep{arandjelovic2013all}, we use the same Hessian-Affine patches
extracted with gravity assumption~\citep{perd2009efficient}. 
Note that this alone results in a $7\%$ gain.

We also compare with global CNN~\citep{babenko2014neural}. 
Our approach outperforms it on Oxford, UKB, and Holidays.
For CNN features with sum-pooling
encoding~\citep{babenko2015aggregating}, we report better results on
Holidays and UKB, and on Oxford with the same evaluation
protocol. Note that their method works better than ours when used
without cropping the queries (58.9\%).

On Holidays, our approach is slightly below the one of 
\cite{gong2014multi}, that uses AlexNet descriptors and VLAD pooling
on large, densely extracted patches. It is however possible to improve
on this result by using the same dimensionality reduction technique
(PCA+whitening) which gives 82.9\% or dense keypoints (82.6\%). 

\begin{table*}[t!]
\begin{center}
  \resizebox{0.7\linewidth}{!}{
    \begin{tabular}{|l|c|c|c|c|c|}\hline
      \bf Method $\backslash$ Dataset & \bf Holidays &\bf UKB &\bf Oxford\\
      \bf VLAD \citep{jegou2012aggregating} & 63.4 & 3.47 & - \\
      \bf VLAD++ \citep{arandjelovic2013all} & 64.6 & - & 55.5* \\
      \bf Global-CNN \citep{babenko2014neural} &  79.3 & 3.56 & 54.5 \\
      \bf MOP-CNN \citep{gong2014multi} & 80.2 & - & -\\ 
      \bf Sum-pooling OxfordNet~\citep{babenko2015aggregating} & 80.2 & 3.65 & 53.1 \\\hline
      \bf Ours & 79.3 ({\bf 82.9})  & \bf 3.76 &  49.8 ({\bf 56.5*}) \\\hline
\end{tabular}}
\end{center}
\caption{\label{tab:imresultsSOA}Comparison with state-of-the-art
  image retrieval results. Results with * use a Hessian-Affine
  detector with gravity assumption~\citep{perd2009efficient}. Our best
  result on Holidays uses whitening as in MOP-CNN.}

\end{table*}

\section{Conclusion}

We showed that Convolutional Kernel Networks (CKNs) offer similar and sometimes even
better performances than classical Convolution Neural Networks (CNNs) in the
context of patch description, and that the good performances observed
in patch retrieval translate into good performances for image retrieval,
reaching state-of-the-art results on several standard benchmarks. 
The main advantage of CKNs compared to CNNs is their very fast training time, and the
fact that unsupervised training suppresses the need for manually
labeled examples. It is still unclear if their success is due to the
particular level of invariance they induce, or their low training time
which allows to efficiently search through the space of
hyperparamaters. We leave this question open and hope to answer it in  
future work. 

{\bf Acknowledgments.} This work was partially supported by projects
``Allegro'' (ERC), ``Titan'' (CNRS-Mastodon), ``Macaron''
(ANR-14-CE23-0003-01), the Moore-Sloan Data Science Environment at NYU
and a Xerox Research Center Europe collaboration contract. We wish to
thank~\cite{fischer2014descriptor, babenko2014neural,gong2014multi}
for their helpful discussions and comments. We gratefully acknowledge the support of NVIDIA Corporation with the donation of the GPUs used for this research.

\bibliographystyle{spbasic} 
\bibliography{biblio}

\end{document}